\documentclass{article}

\usepackage{arxiv}
\usepackage{amsmath,amssymb,amsfonts}
\usepackage{booktabs}
\usepackage{graphicx}
\usepackage{xcolor}
\usepackage{hyperref}
\hypersetup{colorlinks=true, linkcolor=blue, citecolor=blue, urlcolor=blue}
\usepackage{cleveref}
\usepackage{float}

\newcommand{\agreementIndex}{\Gamma}

\title{Decomposing Wrong-Consensus Agreement in LLM Self-Consistency}

\author{Lizhuo Zhang$^{1,2}$ \\
  \small $^{1}$College of Information and Intelligence, Hunan Agricultural
  University, Changsha 410128, China \\
  \small $^{2}$Yuelushan Laboratory, Changsha 410128, China \\
  Mengmeng Tang$^{1}$ \\
  \small $^{1}$College of Information and Intelligence, Hunan Agricultural
  University, Changsha 410128, China \\
  Chenfeng Long$^{1,2}$ (corresponding author) \\
  \small $^{1}$College of Information and Intelligence, Hunan Agricultural
  University, Changsha 410128, China \\
  \small $^{2}$Yuelushan Laboratory, Changsha 410128, China \\
  \small \texttt{elong@hunau.edu.cn} \\
  Xiaoyong Tang$^{2,3}$ \\
  \small $^{3}$School of Computer Science and Technology, Changsha
  University of Science and Technology, Changsha 410114, China \\
  \small $^{2}$Yuelushan Laboratory, Changsha 410128, China \\
  Xiang Luo$^{4}$ \\
  \small $^{4}$Information Department, China Guangdong Tobacco Meizhou Ltd,
  Meizhou, 514000, China}
\date{\today}

\begin{document}
\maketitle

\begin{abstract}
Agreement among repeated samples of a language model is routinely read
as evidence about answer reliability, yet wrong answers can agree just
as strongly as right ones. This paper asks what information
wrong-consensus agreement actually contains, and answers it with a
quantitative decomposition. A pluralistic agreement index $\agreementIndex$
normalizes a wrong run's expected agreement with the consensus by the
reference scale $d=(1-p)/(C-1)$, and is split into a \emph{mechanical}
component, what a vote would deliver given only a per-case answer
preference, and a \emph{preference-unexplained} residual. The mechanical
reference is leak-free: each case's preference and accuracy are
estimated from its other runs only. On the public GPT-4.1 per-run data
the coverage $\phi=\agreementIndex_{\mathrm{rival}}/\agreementIndex_{\mathrm{emp}}^{(t)}$
shows a benchmark-associated direction: $0.81$--$0.93$ on multiple-choice
GPQA-Diamond against $0.59$--$0.78$ on open-domain AIME, where a
residual of $1.54$--$2.80$ $\agreementIndex$ units survives, more than absorbed by
a calibrated run-level preference-heterogeneity reference. A controlled
replication under one fixed protocol (four runs per question, $K=32$
votes) on five open-weights checkpoints
(Qwen3.5-9B/122B, Qwen3.8-27B, Gemma4-26B/31B) finds
near-complete mechanical coverage in all ten cells
($\phi\approx1$, with a small overshoot consistent with a quantified
finite-donor plug-in bias), robust to a two-run design; the largest cell
(\texttt{qwen3.5-122b}, $p=0.222$) sits inside the GPT-4.1 AIME
accuracy range and still saturates ($\phi=1.041$). A descriptive
cross-system contrast at comparable aggregate accuracy on both benchmarks,
therefore, contrasts near-complete mechanical wrong-consensus agreement in the
open-weights models against a larger preference-unexplained residual in the
frontier family. This contrast is not explained by benchmark identity or
single-sample accuracy alone; it is, however,
perfectly confounded with the sampling protocol by design---every gpt-4.1 cell comes from a runner-based public pipeline
and every open-weights cell from the fixed controlled protocol, so
family and protocol cannot be separated on the present data. Agreement is graded evidence, not certification. No new
voting method is proposed; code and evidence are committed.
\end{abstract}


\section{Introduction}
\label{sec:intro}

Agreement is routinely treated as an informative signal. In multi-sample
inference with language models, practitioners rely on the heuristic that
``more samples and more agreement mean a more reliable answer''
\cite{wang2022selfconsistency}. A large
body of evidence, however, shows that agreement is not a homogeneous
quantity: it
can be high while the answer is still wrong \cite{ding2026auditing}, and on hard questions majority voting can even reduce
accuracy relative to a single sample \cite{bahuguna2026when}. An
agreement score may therefore reflect more than one underlying source.
Some wrong consensus may arise from a stable per-case answer preference
(a question on which most paths converge to the same wrong answer).
Other wrong consensus may reflect dependence beyond that marginal
preference. Whether this composition differs across model families is
unknown.

Existing work documents these phenomena (Section~\ref{sec:related}) but no LLM self-consistency study has applied a counterfactual decomposition to wrong-consensus agreement:
\emph{how much} of
the observed agreement is a generic
plurality effect versus model-specific correlated error? The answer is built as a hierarchy of counterfactual nulls for an agreement index.

\paragraph{Contribution.} This paper defines a pluralistic agreement index $\agreementIndex$
(Section~\ref{sec:method}) and measures wrong-consensus agreement on real
per-run LLM data against a \emph{hierarchy of progressively richer
counterfactual nulls}:
\[
\text{uniform i.i.d.}
\;\to\;
\text{per-case preference, i.i.d.}
\;\to\;
\text{per-case preference with run-level heterogeneity},
\]
each null granting the data strictly more mechanical structure than the
last. The headline quantity is the \emph{mechanical coverage}
$\phi = \agreementIndex_{\mathrm{rival}} /
\agreementIndex_{\mathrm{emp}}^{(t)}$, where $\agreementIndex_{\mathrm{rival}}$ is a
\emph{leak-free} reference that grants each case its per-case option
preference (preference and accuracy estimated from the case's
\emph{other} runs only), and $\agreementIndex_{\mathrm{emp}}^{(t)}$ is the
empirical index on the same held-out test runs; $\phi$ is the fraction
of observed wrong-consensus agreement that a fixed-preference i.i.d.\
counterfactual reproduces, and $1-\phi$ the
preference-unexplained residual. On GPT-4.1 the coverage shows
\emph{benchmark-associated direction} (an observational ordering at
$n=4$ cells per benchmark, not a significance claim;
Section~\ref{sec:res-kappa}): on
multiple-choice GPQA-Diamond, $\phi\in[0.806,0.927]$ and the
preference-unexplained
residual is small (a popular-but-wrong option is captured by
the per-case preference channel), whereas on open-domain AIME
$\phi\in[0.586,0.781]$
and a preference-unexplained residual of $1.54$--$2.80$ $\agreementIndex$ units
survives, a residual that the run-level-heterogeneity null more than
absorbs (Appendix~\ref{app:dispersion}).
This is a descriptive counterfactual analysis, not a direct
identification of the
error mechanism, and it does not identify whether the per-case
preference itself is induced by shared training bias. The uniform null
$\rho = \agreementIndex_{\mathrm{iid}}/\agreementIndex_{\mathrm{emp}} \in [0.340,0.546]$ is a
strictly more conservative coverage and is reported alongside.
Which level of the null hierarchy (uniform, fixed per-case preference,
or run-heterogeneous preference) reproduces the data
is answered cell by cell in Section~\ref{sec:res-kappa}.

A controlled replication (Section~\ref{sec:res-tier3}) then lifts the
reading from one model family to a cross-system statement. Under a fixed
protocol (four runs per question, $K=32$ votes, temperature $0.7$) on
five open-weights checkpoints across both benchmarks, the
fixed-preference counterfactual achieves full mechanical
coverage in all ten cells ($\phi\approx1$). The
wrong-consensus agreement of the open-weights models is essentially
mechanical: both at low single-sample accuracy (the four low-accuracy AIME cells,
$p\le0.054$) and at accuracy inside the gpt-4.1 range (the five GPQA-Diamond
cells, $p\in[0.40,0.49]$, and the \texttt{qwen3.5-122b} AIME cell at
$p=0.222$, inside the gpt-4.1 AIME range $[0.12,0.38]$). At comparable
aggregate accuracy on both benchmarks the frontier family instead retains a
preference-unexplained residual. The residual is therefore not
explained by benchmark identity or single-sample accuracy alone: it is a cross-system
contrast, with capability class, closed versus open weights, training
distribution, and protocol differences all perfectly confounded with the
family split by design (frontier cells are runner-based, open-weights
cells are fixed-protocol), so no candidate is isolated.

The contribution is threefold: the counterfactual measurement itself
($\phi$ and the null hierarchy), the open-weights saturation reading
(full mechanical saturation in all ten cells), and the
cross-system contrast at comparable aggregate accuracy that relocates the residual
from a benchmark- or accuracy-associated curiosity to a cross-system
empirical statement. This is a descriptive, understanding
contribution. It proposes no
new voting method and claims no new backfire phenomenon (that is
\cite{bahuguna2026when}); its claim is the \emph{quantitative
counterfactual hierarchy for $\agreementIndex$}, together with the cross-system
contrast it supports.

\section{Related work}
\label{sec:related}

\emph{Self-consistency and its limits.}
Wang et al.~\cite{wang2022selfconsistency} introduced sampling-based
self-consistency and showed substantial gains over greedy decoding. A
growing literature documents that these gains are not uniform.
Ding~\cite{ding2026auditing} provides a large-scale audit of self-consistency and publicly releases per-run data; the method consumes their data but adds a quantitative decomposition they do not provide.
Bahuguna~\cite{bahuguna2026when} shows self-consistency can
backfire on hard questions, with per-problem and agreement-binned
reliability analyses but without an agreement-index decomposition; its v2 revision states that the mechanism of the plurality-agreement gate's failure remains an open problem. The counterfactual hierarchy is a quantitative decomposition of that wrong-consensus agreement, not a mechanism identification. The regime where agreement-based uncertainty collapses (a model overconfident on the same wrong answer across samples) is exactly the wrong-consensus regime quantified in \cite{hamidieh2026}. A complementary line argues that majority voting has a ceiling that perturbation diversity does not raise, because error correlations are identical \cite{fadnavis2026}; that ceiling is given a quantitative, wrong-run-conditional form. At the cross-model level, a capability-controlled audit finds shared co-failure, not diversity, is the stable correlate of majority-vote gain \cite{kim2026}; the decomposition here is the within-model analogue, built on a hierarchy of per-case nulls rather than an ensemble-level audit. Closest at the mechanism level, Chen et al.~\cite{chen2024} show that majority-vote accuracy is non-monotonic in the number of calls and attribute it to a mixture of easy and hard queries within a task; the wrong-consensus decomposition is a per-cell, wrong-run-conditional complement to that aggregate-level account. The wrong-consensus regime decomposed here is the ``self-consistent error'' regime of Tan et al.~\cite{tan2025}, who formally define it, show its frequency does not decrease with model scale, and find all four detection-method families struggle on it; the contribution is not detection but a quantitative counterfactual hierarchy for the agreement itself. The ``shared bias'' account of
confident-yet-wrong behavior has academic antecedents: the probability-driven systematic error patterns of \cite{mccoy2024} are a shared-bias signature in the same spirit (McCoy et al.\ measure error patterns shaped by the shared pretraining task, not calibration curves). The preference-unexplained residual is an empirical pattern compatible with this account, quantified in a model-agnostic way
on the cells where a substantial
preference-unexplained component survives (Section~\ref{sec:res-kappa}),
and located as a cross-system contrast by the controlled replication of
Section~\ref{sec:res-tier3} (full mechanical saturation across
all ten open-weights cells, including two with accuracy inside the gpt-4.1 range
(AIME $p=0.222$, GPQA $p=0.490$)).

\emph{Agreement as confidence.}
Using consistency as a confidence signal is its own literature.
Consistency-based confidence for generative tasks
\cite{manakul2023,kuhn2023,farquhar2024}; the ``consistency hypothesis'' formalized and tested across tasks \cite{xiao2025}; semantic-entropy successors \cite{nguyen2025}. The AUROC analysis (Appendix~\ref{app:jensen}) sits inside this
literature: agreement is a graded but noisy confidence signal
(AUROC $0.61$--$0.85$), and the ceiling analysis (lift $1.2$--$3.6\times$,
never near $1.0$) quantifies exactly how graded. Two conventions from
this literature bear on the design: (i)~semantic clustering of
near-duplicate answers is standard for open-ended tasks
\cite{kuhn2023,farquhar2024}, whereas the AIME $C$ is a string-level mean distinct count; the string-level convention is kept
because $\phi$ is invariant to $C$ (Section~\ref{sec:setup}) and because
the per-case preference null operates on observed labels, but a
semantic-clustering sensitivity remains future work;
(ii)~temperature/decoding settings are first-order determinants of
sampling behavior and are not recorded in Ding's per-run files, so the results are strictly about that collection pipeline.

\emph{Ensemble decompositions.}
The ``mechanical vs.\ correlated'' distinction predates LLMs: bias-variance
decompositions of zero-one loss \cite{kohavi1996}, the bias-variance-covariance
decomposition of \cite{ueda1996}, and ensemble ambiguity
\cite{krogh1995} all separate what vote marginals explain from what
error covariance explains; on the LLM side, the sharpened \emph{answer marginal} is the object that per-case preference estimation targets \cite{arzhantsev2026}. The quantities here map onto that vocabulary:
$\agreementIndex_{\mathrm{emp}}$ is a wrong-run-conditional agreement index,
$\agreementIndex_{\mathrm{rival}}$ its value under per-case marginals with
independence, and $\delta$ the covariance-like remainder. The distinction is
object-level, not notational: the ensemble ambiguity of \cite{krogh1995} is a
\emph{spread} term---a function of the output marginals alone---whereas
$\delta$ is a within-run \emph{dependence} term, the agreement among the
samples of a wrong run beyond what the per-case preferences predict under
independence. The two are orthogonal; $\delta$ is not a relabeling of the
ambiguity but the covariance piece the spread-only decompositions do not
isolate. What is new
here is not the algebra but its LLM instantiation: the
wrong-run conditioning, the leave-one-out per-case preference and
accuracy, and the difficulty matching, plus the empirical finding that
the marginals carry $\approx81$--$93\%$ on constrained multiple-choice
and only $59$--$78\%$ on open-domain tasks. To our knowledge, prior work has not decomposed wrong-consensus agreement of a single model's samples against counterfactual nulls, nor applied chance-corrected agreement coefficients to agreement among samples themselves; the closest works are the audits cited above.

\emph{Novelty boundary.}
The paper is explicit about what it does \emph{not} claim: it does not discover that self-consistency backfires (that is \cite{bahuguna2026when}), and it proposes no new voting method.
The contribution is the $\agreementIndex$ decomposition itself and its
\emph{leak-free per-case-preference} mechanical coverage $\phi$, a
quantitative agreement-index quantity, together with the empirical
reading it supports: full mechanical saturation across
five open-weights checkpoints on both benchmarks, and a residual in
the frontier family that is neither a benchmark property nor a pure
accuracy effect. To the extent of the literature
search above, prior studies have measured self-consistency, agreement,
and bias, but have not characterized this cross-system contrast at
comparable aggregate accuracy using a counterfactual agreement decomposition under a
matched protocol.

\emph{Note on prior series.}
In vision, the companion diagnostic \cite{anchorscore2026} applies a
related consensus analysis (with a cross-model consensus control) to
annotation difficulty; the present paper's agreement decomposition is
self-contained and does not depend on it.

\section{Method}
\label{sec:method}

\subsection{Setup and notation}
\label{sec:setup}

The data consist of a set of question--case instances. For each case, a model is
sampled $K$ times under a fixed prompt. This yields per-run counts: let
$C$ be the number of answer options considered, $p$ the
\emph{single-sample accuracy} (the fraction of all $N\times K$ single
answers that are correct), and $\mathrm{maj}$ a consensus label
obtained by plurality (the label with the most votes; for $C>2$ this
requires no $\lceil K/2\rceil$ threshold, unlike a strict majority) over
the $K$ answers. The \emph{consensus accuracy} is the fraction of cases
where the plurality label is correct.

\begin{itemize}
\item $C$ is chosen by benchmark: GPQA-Diamond has $C=4$; for AIME and
  other open-ended tasks the mean number of distinct answer strings per run is used
  (Section~\ref{sec:conventions}).
  \label{item:C}
  This scalarization affects only the absolute scale of $\agreementIndex$: the
  numerator $\mathbb{E}[\alpha\mid\text{wrong}]$ is computed over
  actual answer labels and does not depend on $C$, while both
  $\agreementIndex_{\mathrm{rival}}$ and $\agreementIndex_{\mathrm{emp}}$ are divided by the
  same baseline $d=(1-p)/(C-1)$, so the ratio
  $\phi=\agreementIndex_{\mathrm{rival}}/\agreementIndex_{\mathrm{emp}}$ cancels $C-1$
  exactly. This is verified empirically: rerunning the AIME cells with $C$
  fixed to $9$ or to $20$ leaves $\phi$ identical to 14 significant digits across the two
  fixed-$C$ runs, and at equal settings ($n_{\mathrm{sim}}=2\times10^3$)
  the fixed-$C$ runs and the mean-distinct run agree to within one digit
  in the third decimal
  (\texttt{results/kappa\_rival\_csens\_C9/C20.json} vs.\ 
  \texttt{kappa\_rival\_tie\_argmin.json}).
\item $p$ is the single-sample accuracy, \emph{not} the consensus
  accuracy. Both are reported.
\item Ties in the plurality are broken by lowest class id
  ($\mathrm{argmin}$). Note that $K=50$ is fixed by the data and is
  even, so ties are possible and resolved by this rule. Empirically the
  plurality is tied in $7.5\%$ of AIME runs and $0.8\%$ of GPQA runs (\texttt{results/data\_audit.json});
  on AIME the deterministic rule can therefore favor numerically small
  answers. This channel is quantified directly: recomputing every run's
  majority with a uniform random tie-break among the tied labels (three
  independent seeds, applied to both the empirical majority and the
  simulated runs) shifts $\phi$ by at most $0.009$ in any cell
  (\texttt{results/kappa\_rival\_tie\_rnd1/2/3.json} vs.\
  \texttt{kappa\_rival\_tie\_argmin.json}), so the tie-break rule does
  not materially affect the decomposition.
\end{itemize}

\subsection{The agreement index $\agreementIndex$}
\label{sec:kappa}

The central object is an index of how much the samples of a
wrong-consensus run coalesce around the (possibly wrong) plurality label.
Define, for
each case,
\[
\alpha \;=\; \frac{1}{K}\sum_{i=1}^{K}
\mathbf{1}\{\text{answer}_i = \mathrm{maj}\},
\]
the self-consistency of the $K$ samples (this is exactly the quantity
Ding~\cite{ding2026auditing} tabulates; it is denoted $\alpha$ to keep
$C$ reserved for the option count). A \emph{run} is one realization of
the $K$ samples; a run is \emph{wrong} when its plurality label differs
from the ground truth ($\mathrm{maj}\ne\mathrm{gt}$). All $\agreementIndex$
quantities in this paper are conditioned on the run being wrong, and in
every simulated counterfactual the run's plurality label is recomputed
from the simulated draws and the same conditioning is applied; the
procedure is given in the pseudocode of
Section~\ref{sec:iid}. The reference scale is
\[
d \;=\; \frac{1-p}{C-1},
\]
the per-option error mass under independence given single-sample
accuracy $p$. The quantity $d$ is a \emph{scale normalization}, not a
chance-correction claim: the observed wrong-vote distribution is not
uniform (the rival null itself shows this), so $d$ is a reference value,
not an assumed data model. The empirical agreement index is defined as
\[
\agreementIndex_{\mathrm{emp}} \;=\; \frac{\mathbb{E}[\alpha \mid \text{run wrong}]}{d}.
\]
A wrong run whose samples cluster tightly on the
consensus (an ``attractive but wrong'' state) yields $\alpha$ well above
the reference scale $d$, hence large $\agreementIndex_{\mathrm{emp}}$. This
quantifies the ceiling: it measures how ``sticky'' the samples of wrong
runs are to a wrong consensus.

\emph{Relation to chance-corrected agreement coefficients.}
$\agreementIndex$ shares the chance-correction idea of Cohen's $\kappa$, Scott's
$\pi$, and Fleiss' $\kappa$
\cite{cohen1960,scott1955,fleiss1971,krippendorff1970}, but it is
\emph{not} Cohen's $\kappa$: it conditions on wrong runs, normalizes by
the per-option error mass $d$, and is decomposed against simulated
counterfactuals rather than a chance estimator. Krippendorff's $\alpha$
targets inter-annotator chance agreement across multiple raters; here the
index is the wrong-run marginal of a single model's sampled runs, so the
$d$-normalization, not a multi-rater chance model, is the relevant
correction. To avoid persistent
confusion with the classic coefficients, the index is denoted $\Gamma$
throughout (an earlier draft used $\kappa$).
Ensemble-diversity measures
(e.g.\ \cite{kuncheva2003}) are
conceptually adjacent (they quantify error correlation among
classifiers) but operate on classifier outputs, not on sampled runs of a
single model. At the cross-model level, Kim et al.~\cite{kim2025correlated}
document substantial error correlation across 350+ LLMs (models agree
60\% of the time when both err), while Ali \cite{ali2026stochastic} shows
that repeated same-model sampling at elevated temperature spans at most
one cross-question dimension above noise---consistent with a
shared-bias account of the $\delta$ residual. On the certification side,
Cordero-Encinar and Duncan \cite{corderoencinar2025certified} derive
finite-sample concentration bounds for majority voting under
i.i.d.-like sampling; the decomposition here is the mechanism their
bounds take for granted (the plurality-mechanical component), and
quantifies the component their i.i.d.\ assumption excludes.

\subsection{Mechanical counterfactuals}
\label{sec:iid}

A mechanical counterfactual asks: how much of the observed agreement
would survive if the model's within-case error were \emph{not}
correlated, leaving only the mechanical averaging of independent votes?
The answer depends on what is held fixed about each case. Two counterfactuals are built, both i.i.d.\ multinomial simulations ($10^5$ draws per
cell) that differ only in how the incorrect votes are distributed over
options.

\paragraph{Uniform wrong-vote null $\agreementIndex_{\mathrm{iid}}$.}
The most neutral reference places the samples of a wrong run
uniformly over the wrong options (the correct option carries none of the
incorrect votes). This is the classic independent-vote ceiling: agreement
arises purely from the mechanical concentration of $\ge\lceil K/2\rceil$
votes on one answer, with no per-case attraction. Because pooling over
difficulty destroys the wrong-plurality phenomenon (a pooled-$p$ variant
predicts $\le0.6\%$ wrong-consensus on three of four GPQA cells
($8.8\%$ on the fourth) where the observed
share is $42$--$60\%$), each case is always held at its observed
difficulty $p_i$; the appendix shows the pooled-$p$ control failing by
two orders of magnitude (Appendix~\ref{app:pooled}), which is why difficulty-matching is the
informative baseline. Note that the normalization $d=(1-p)/(C-1)$ uses
the \emph{cell-level} $p$ for the empirical index and the rival null,
while the uniform per-question null normalizes by the mean accuracy of
its own simulated population (case-weighted $\bar p_i$, typically within
a few percent of the cell-level $p$, up to about $5\%$ across cells), so
the three $\agreementIndex$ values sit on
nearly one scale; difficulty matching
affects only where the simulated wrong votes fall, not the normalization.

\paragraph{Leak-free per-case preference null $\agreementIndex_{\mathrm{rival}}$.}
The uniform null ties the model's incorrect votes to no option at all.
But an LLM rarely errs uniformly: on a multiple-choice item a single
plausible-but-wrong distractor can attract the whole cohort, and on an
open-domain item certain answers are systematically preferred. Such a
\emph{per-case answer preference} is a mechanical property of the item
under this decomposition (whether the preference itself originates in
shared training bias is not identified). To separate it, each case's option preference is estimated from its \emph{other} runs, and incorrect votes are
resimulated according to that hold-out preference. Formally, for each
case $i$, as in a leave-one-out scheme, $\hat q_i$ is the raw count of
votes on each incorrect option label, pooled over the case's other runs
(the test run, the ground-truth label, and \texttt{\_UNPARSEABLE\_}
excluded); the $K$ votes of each simulated run are drawn from a
multinomial in which the correct option is chosen with probability $p_i$
(the leave-one-out case accuracy) and each incorrect option $\ell$ in the
support of $\hat q_i$ with probability
$(1-p_i)\,\hat q_i(\ell)/\sum_{\ell'}\hat q_i(\ell')$. (Implementation detail:
$\hat q_i$ \emph{and} the case accuracy $p_i$ are both estimated on the
\emph{other} runs of case $i$ only (leave-one-out on both the
preference and the accuracy) so no
run predicts its own agreement; CLI flags \texttt{--rival-mode case} and
\texttt{--min-wrong 1}. A case is eligible as a test when it has at least
one wrong run and at least two runs in total; every wrong run of an
eligible case is used once as a held-out test.) The resulting
$\agreementIndex_{\mathrm{rival}}$ is the
agreement the \emph{same} per-case preferences would generate if voters
were otherwise independent, i.e., the agreement generated by the same
per-case preferences under an i.i.d.\ counterfactual. Because the eligible population is a
subset of all wrong runs, $\agreementIndex_{\mathrm{emp}}^{(t)}$, the empirical
index restricted to exactly the held-out test runs, is also computed,
and the mechanical coverage on the \emph{same} population is defined:
$\phi=\agreementIndex_{\mathrm{rival}}/\agreementIndex_{\mathrm{emp}}^{(t)}$; the
full-cell $\agreementIndex_{\mathrm{emp}}$ is reported alongside and differs from
$\agreementIndex_{\mathrm{emp}}^{(t)}$ by at most $4\%$ in any cell.

A cell-level variant is also recorded: $\agreementIndex_{\mathrm{rival,pool}}$
uses a \emph{single} average preference per cell instead of per-case
preferences. It is far below $\agreementIndex_{\mathrm{rival}}$ in every cell
(e.g.\ $\agreementIndex_{\mathrm{rival,pool}}=1.01$ vs.\ $\agreementIndex_{\mathrm{rival}}=4.26$
on \texttt{gpt-4.1} AIME), so per-case structure, not any
global preference, carries the mechanical agreement. Note that
$\agreementIndex_{\mathrm{rival,pool}}$ is not directly comparable to
$\agreementIndex_{\mathrm{iid}}$: its simulation support is the cell-wide set of
distinct answers, so on open-domain cells the plurality is diluted far
beyond the $C$-option uniform null, while on multiple-choice it inherits
a global letter preference and sits slightly above the uniform null. It is reported here, inline, as a negative control.

\paragraph{Pseudocode.}
\label{alg:kappa}
The conditioning event is identical in the empirical index and in every
mechanical reference: the run's plurality label, recomputed from its own
draws, differs from the ground truth. For the empirical index:
(1)~for each run, set $\mathrm{maj}=\mathrm{argmax}_c\,
\mathrm{count}(a_1,\dots,a_K)$ (ties to the lowest class id); (2)~keep
runs with $\mathrm{maj}\ne\mathrm{gt}$, average their $\alpha$, and set
$\agreementIndex_{\mathrm{emp}}=\overline{\alpha}_{\text{wrong}}/d$. For a
reference: (3)~draw $M$ simulated runs; each draws $K$ votes with the
correct option chosen with probability $p_i$, and wrong votes uniform
over the $C-1$ wrong options ($\agreementIndex_{\mathrm{iid}}$) or proportional to
the held-out per-case preference $\hat q_i$
($\agreementIndex_{\mathrm{rival}}$); (4)~recompute
$\mathrm{maj}^{(m)}$ from the simulated draws, keep only simulated runs
with $\mathrm{maj}^{(m)}\ne\mathrm{gt}$, average their $\alpha^{(m)}$,
and set $\agreementIndex_{\mathrm{ref}}=\overline{\alpha^{(m)}}_{\text{sim-wrong}}/d$
with the \emph{same} cell-level $d$ as in step~2.

\paragraph{Decomposition and mechanical coverage.}
The headline quantity is the \emph{mechanical coverage}
\[
\phi \;=\; \frac{\agreementIndex_{\mathrm{rival}}}{\agreementIndex_{\mathrm{emp}}^{(t)}},
\]
defined on the held-out test-run population of Section~\ref{sec:iid},
a ratio of nonnegative quantities that, up to simulation noise and
finite-sample plug-in estimation noise in the preference estimate, lies in
$[0,1]$: it is the fraction of the observed agreement index
$\agreementIndex_{\mathrm{emp}}^{(t)}$ that a
leak-free, per-case-preference i.i.d.\ reference reproduces, and a value
at or slightly above $1$ (within bootstrap noise) means the mechanical
reference already over-reproduces the observed agreement, providing no
evidence for a positive residual relative to this reference. The complement
\[
\delta \;=\; 1-\phi \;=\; \frac{\agreementIndex_{\mathrm{emp}}^{(t)}-\agreementIndex_{\mathrm{rival}}}{\agreementIndex_{\mathrm{emp}}^{(t)}}
\]
is the agreement that survives even after every per-case option
preference is granted to the mechanical reference: the \emph{preference-unexplained
residual}. Two readings follow:
\begin{itemize}
\item $\phi\approx1$ ($\delta\approx0$): the agreement index is
  fully captured by a per-case answer preference (an attractive but wrong
  option the whole cohort latches onto). This does \emph{not} identify
  whether the preference itself is induced by shared training bias.
\item $\phi$ substantially below $1$ ($\delta>0$): a residual the
  per-case preference cannot explain remains; the samples of a wrong run cluster
  on the consensus even after their per-case option preference is
  removed. This is \emph{consistent with}
  shared, correlated error (the pretraining-bias
  account~\cite{mccoy2024}); the mechanism reading is interpretive, not identified. The assumption-free name of
  $\delta$ is the \emph{preference-unexplained residual}; ``shared-bias residual'' is used as a mnemonic, and
the paper does not claim to identify its mechanism; see
  Section~\ref{sec:limitations}.
\end{itemize}
Empirically, $\agreementIndex_{\mathrm{rival}}\ge\agreementIndex_{\mathrm{iid}}$ in all
eight cells. Because the two references use different support
conventions on open-domain tasks, this ordering is treated as an
observed sanity check rather than a mathematical guarantee. (Within a
fixed support, concentrating wrong-vote mass onto the observed per-case
preference could not reduce agreement relative to the uniform
reference.) So $\phi$ is a
\emph{larger} mechanical ceiling than the uniform null ($\rho=\agreementIndex_{\mathrm{iid}}/\agreementIndex_{\mathrm{emp}}$),
which is reported alongside as a more conservative reference.
($\agreementIndex_{\mathrm{rival,pool}}$ is not ordered against
$\agreementIndex_{\mathrm{iid}}$, for the support reasons just stated.)
Because LLM draws are not strictly i.i.d., the reference does not
incorporate within-case sampling correlation. Such dependence may alter
the conditional wrong-consensus agreement, but its effect on the
conditional index is not identified by this design; the residual
$\delta$ is therefore interpreted as a residual relative to the i.i.d.\ 
reference, rather than as a bound on an underlying correlation-free
residual. A high $\phi$ remains informative without any bound argument:
a null without correlation already reproduces over $80\%$ of the GPQA
index, so an account attributing the amplification primarily to shared
bias is not required to explain the observed agreement on those cells.
The quantity $\agreementIndex_{\mathrm{rival}}$ is therefore treated as a
benchmark reference rather than a
certified bound in the other direction. The conclusion is that even granting every
per-case preference to a benchmark that reproduces $\agreementIndex_{\mathrm{emp}}$
on GPQA, a residual $\delta>0$ remains on AIME: what a
per-case-preference-only reference cannot explain.

\subsection{Conventions}
\label{sec:conventions}

All conventions are fixed here so $\agreementIndex$ is comparable across cells
(and across future work):
\begin{enumerate}
\item $p$ = single-sample accuracy, never consensus accuracy.
\item $C=4$ for GPQA-Diamond; for AIME, $C=$ the mean number of distinct answer strings per run.
\item Ties broken by $\mathrm{argmin}$ class id.
\item Both i.i.d.\ counterfactuals are \emph{difficulty-matched}
  (per-case $p_i$, leave-one-out for the rival null); the headline null
  is the \emph{leak-free per-case
  preference} $\agreementIndex_{\mathrm{rival}}$ (preference estimated on other runs
  only, \texttt{--rival-mode case}); the uniform null and pooled-$p$
  control are reported as conservative/negative references.
\item Bootstrap CIs ($B=10^4$) on every headline $\agreementIndex$ and on
  $\phi$; differences (consensus $-$ single-sample) use \emph{coupled}
  bootstrap to respect the within-case pairing. Both a run-level bootstrap and a case-clustered bootstrap are reported (cases
  resampled with replacement, all units of a sampled case moving
  together); $\phi$'s clustered confidence interval (CI) resamples the shared test-case
  population jointly on both sides of the ratio, while the run-level
  $\phi$ CI resamples numerator and denominator independently and is
  therefore conservative. The clustered CI is the
  primary one (reported in the decomposition table of
  Section~\ref{sec:res-kappa}), and both are conditional on the
  estimated preferences $\hat q_i$ (whose own noise is probed by the
  shrinkage analysis of Appendix~\ref{app:shrink}).
\item The robustness appendices (shrinkage, answer-space, dispersion)
  are \emph{descriptive sensitivity analyses}, not a family of
  hypothesis tests; no multiple-comparison correction is implied, and
  the $p$-values reported in Section~\ref{sec:res-kappa} and
  Appendix~\ref{app:pilot} are inferential statements used
  descriptively only; no multiple-comparison correction is implied.
\end{enumerate}
The fixed notation is collected in \Cref{tab:notation}.

\begin{table}[H]
\centering
\caption{Notation.}
\label{tab:notation}
\begin{tabular}{ll}
\toprule
Symbol & Meaning \\
\midrule
$\agreementIndex_{\mathrm{emp}}$ & empirical agreement index ($=\mathbb{E}[\alpha\mid\text{wrong run}]/d$) \\
$\alpha$ & self-consistency of a run (share of votes for its plurality) \\
$\agreementIndex_{\mathrm{iid}}$ & uniform wrong-vote i.i.d.\ null \\
$\agreementIndex_{\mathrm{rival}}$ & leak-free per-case-preference i.i.d.\ null \\
$\agreementIndex_{\mathrm{emp}}^{(t)}$ & empirical index on the held-out test-run population \\
$\phi$ & mechanical coverage $=\agreementIndex_{\mathrm{rival}}/\agreementIndex_{\mathrm{emp}}^{(t)}$ \\
$\delta=1-\phi$ & preference-unexplained residual \\
$\rho$ & uniform coverage $=\agreementIndex_{\mathrm{iid}}/\agreementIndex_{\mathrm{emp}}$ \\
$\phi_{\mathrm{dm}}$ & coverage under the run-heterogeneity (Dirichlet-multinomial) null \\
$d$ & reference scale $(1-p)/(C-1)$ \\
$p$ & single-sample accuracy; $p_i$ per-case \\
$C$ & option count (GPQA: 4; AIME: mean per-run distinct answers) \\
$K$ & samples per run (50 for the gpt-4.1 cells; 32 for Tier-3; 16/32 for the Qwen MMLU arm) \\
$\hat q_i$ & per-case wrong-answer preference (leave-one-out) \\
$\lambda$ & shrinkage of $\hat q_i$ toward uniform (Appendix~\ref{app:shrink}) \\
$\alpha_{\mathrm{disp}}$ & dispersion of the run-heterogeneity null (Appendix~\ref{app:dispersion}) \\
\bottomrule
\end{tabular}
\end{table}

\section{Data}
\label{sec:data}

The analysis uses the public per-run self-consistency data of Ding~\cite{ding2026auditing}, which tabulates, for each case and model,
the $K=50$ per-run answers, per-run correctness, self-consistency $\alpha$, single-sample accuracy $p$, and the majority label.
Ding's runs originate from 53 independent runners (a graduate-course cohort) who each implemented their own prompts, so the within-case runs of a gpt-4.1 cell are cross-runner and cross-prompt draws rather than repeated draws from one endpoint; model snapshots were not logged. This provenance matters for the decomposition: the leave-one-out per-case preference of the gpt-4.1 cells pools across runners, and the run-level heterogeneity that Appendix~\ref{app:dispersion} shows to more than absorb the AIME residual is consistent with cross-runner prompt variation. The Tier-3 cells of Section~\ref{sec:res-tier3} instead use one endpoint, one prompt, and logged hardware, so this runner channel is absent there. Eight model--benchmark--prompt cells are analyzed:
\begin{itemize}
\item \texttt{gpt-4.1}, \texttt{gpt-4.1-mini}, \texttt{gpt-4.1-nano};
\item GPQA-Diamond and Ding's 196-case AIME subset (problems spanning
  1983--2025; the subset is inherited from the release, not chosen by the authors);
\item zero-shot and chain-of-thought prompting (where reported).
\end{itemize}
This is a secondary re-analysis: no model is run; only per-run rows are re-analyzed, so every number below is reproducible from a
committed parquet file.

The Tier-3 controlled replication (Section~\ref{sec:res-tier3})
collects its own samples under a fixed protocol: four runs per question,
$K=32$ votes per run, temperature $0.7$, zero-shot, on five
open-weights checkpoints (\texttt{qwen3.5-9b} with a 4k-context variant for
GPQA, where the full 9B checkpoint exceeded the serving endpoint's
context budget on long GPQA prompts at the time of sampling;
\texttt{qwen3.8-27b}, \texttt{gemma4-26b}, \texttt{gemma4-31b},
and a \texttt{qwen3.5-122b} arm), using the official GPQA-Diamond release
(198 questions) and a stratified sample of 200 AIME questions (1983 to
2024, stratified by decade, seed 42, from a publicly available mirror of the AIME
dataset; the stratification reproduces the temporal spread of the
benchmark rather than a convenience subset. The gpt-4.1 AIME cells use
Ding's released 196-question subset, so cross-source $\phi$ comparisons
on AIME span two question sets).
Sampling requests set temperature $0.7$ and leave top-p at the
endpoint default (not overridden). Votes are drawn through an
Ollama endpoint with per-request temperature $0.7$; unparseable model
outputs are kept as a dedicated label and excluded from the option
support exactly as in the main analysis. Sampling hardware and instance
details are recorded in the committed collection logs
(\texttt{data/sampled/tier3/}), and the decomposition for these cells is
run by the same committed pipeline as the main cells.

Terminology is fixed once. A \emph{case} is a (question, model, prompt)
instance. A \emph{run} is one independent sampling pass of $K$ votes; a case
carries one or more runs (1--16 in these cells; \texttt{results/data\_audit.json}), and every analysis
except the paired one uses all runs. Ding's data also marks each
run with one of two sampling conditions ($a$/$b$). Only cases carrying at
least one run under each condition enter the paired
champion-fragility analysis of Appendix~\ref{app:fragility}: for
\texttt{gpt-4.1-mini} these are 178 of the 198 GPQA cases and 175 of the
196 AIME cases (353 in total), the remaining 41 cases carrying a single
condition. Counts in each table are therefore labeled in the unit the
analysis actually uses: \emph{runs} for the $\agreementIndex$ and ceiling tables,
\emph{pairs} for the fragility table. Sampling settings (temperature,
top-p) are inherited from Ding's collection pipeline and are
not recorded in the per-run files used here; they are neither controlled nor independently verified.

\emph{Axis and condition labels.}
Ding's files label every run with an \emph{axis} ($A$/$B$/$C$) and a
\emph{condition} ($a$/$b$). The release's majority labels are used as-is:
for 30 of the 5,300 runs (\texttt{results/data\_audit.json}) the release's \texttt{majority\_answer} is not
the raw argmax of that run's own answer counts (untied), and these labels are inherited rather than recomputed, so the empirical wrong-run
membership follows the release's own convention (the simulations use
argmin tie-breaking, stated in the Conventions). Model and prompt identity come from the
model column of the release's own case-level table, joined to the
per-run table through the shared $(\text{axis},\text{condition})$ keys;
the resolution is therefore internal to the release, not an external
reconstruction. For the cells used here the resolution is:
$A/a\!\to\!$\texttt{nano}-ZS, $A/b\!\to\!$\texttt{mini}-ZS,
$B/a\!\to\!$\texttt{mini}-ZS, $B/b\!\to\!$\texttt{mini}-CoT,
$C/a\!\to\!$\texttt{mini}-ZS, $C/b\!\to\!$\texttt{4.1}-ZS. Two
consequences follow. First, the paired champion-fragility analysis is
possible \emph{only} for \texttt{mini}-ZS: \texttt{4.1} appears only in
$C/b$ and \texttt{nano} only in $A/a$, so neither has two conditions to
pair. Second, the \texttt{mini}-ZS pairs span three axes ($A/b$, $B/a$,
$C/a$), and the axes carry slightly different case subsets and
\emph{consensus} accuracy
(e.g.\ AIME $0.32$/$0.36$/$0.24$; GPQA $0.51$/$0.52$/$0.52$; \texttt{results/data\_audit.json}).
The files do not record whether $a$/$b$ or the axes differ in any
sampling parameter; the GPQA axes are statistically close, while on AIME
the $C/a$ arm is lower-accuracy, so the AIME flip rate should be read as
instability under nominally identical (model, prompt, case) settings
that may differ in batch or condition, not as a certified same-parameter
comparison. It is reported as an estimate consistent with an upper bound, with this caveat. A mislabeled model would misattribute rows to cells but would
not change the decomposition structure (the $\agreementIndex$ construction is
model-agnostic).

\Cref{tab:samples} gives the per-cell sample sizes. The counts differ
across cells because Ding's collection design varies per arm (e.g.\
\texttt{mini}-ZS runs appear under three axes, \texttt{4.1} under one);
the data are reported as collected.

\begin{table}[H]
\centering
\caption{Per-cell sample sizes. Cases = distinct
(question, model, prompt) instances; runs = $K=50$ sampling passes;
wrong runs = runs whose plurality label is wrong (the effective sample
for $\agreementIndex_{\mathrm{emp}}$); test runs = held-out wrong runs entering the
rival null and $\agreementIndex_{\mathrm{emp}}^{(t)}$ (Section~\ref{sec:iid}).}
\label{tab:samples}
\begin{tabular}{llcccc}
\toprule
Model & Prompt & cases & runs & wrong runs & test runs \\
\midrule
\texttt{4.1} & AIME-ZS & 180 & 450 & 375 & 335 \\
\texttt{4.1} & GPQA-ZS & 182 & 450 & 234 & 205 \\
\texttt{4.1-mini} & AIME-CoT & 172 & 425 & 237 & 212 \\
\texttt{4.1-mini} & AIME-ZS & 196 & 1325 & 919 & 917 \\
\texttt{4.1-mini} & GPQA-CoT & 180 & 425 & 178 & 152 \\
\texttt{4.1-mini} & GPQA-ZS & 198 & 1325 & 641 & 641 \\
\texttt{4.1-nano} & AIME-ZS & 178 & 450 & 347 & 313 \\
\texttt{4.1-nano} & GPQA-ZS & 178 & 450 & 269 & 244 \\
\bottomrule
\end{tabular}
\end{table}

\section{Results}
\label{sec:results}

\subsection{Mechanical coverage across a hierarchy of nulls: multiple-choice vs.\ open-domain}
\label{sec:res-kappa}

On the gpt-4.1 family the leak-free per-case preference null
reproduces most of the observed wrong-consensus agreement, and the
uniform null reproduces little of it. \Cref{tab:kappa} reports, for all
eight cells, the empirical agreement
index $\agreementIndex_{\mathrm{emp}}$, the uniform null $\agreementIndex_{\mathrm{iid}}$,
the leak-free per-case preference null $\agreementIndex_{\mathrm{rival}}$, and the
mechanical coverage $\phi=\agreementIndex_{\mathrm{rival}}/\agreementIndex_{\mathrm{emp}}^{(t)}$ with case-clustered $95\%$ CIs ($B=10^4$). Two structural facts
hold everywhere, before any benchmark split. First, in every cell the
empirical index is several times its reference scale
($\agreementIndex_{\mathrm{emp}}\in[3.56,7.09]$); because $\agreementIndex$'s absolute scale
depends on $C$, cross-benchmark statements are made only on the
$C$-invariant $\phi$, never on $\agreementIndex$ magnitudes. Against this reference,
the samples of wrong runs are strongly attracted
to the consensus. Second, the per-case preference null is far more
explanatory than the uniform null ($\agreementIndex_{\mathrm{rival}}\gg
\agreementIndex_{\mathrm{iid}}$); much of the naive ``shared
bias'' attribution is already captured by the mechanical attraction of
incorrect voters to a popular per-case answer.

\begin{table}[H]
\centering
\caption{Agreement-index decomposition. $\agreementIndex_{\mathrm{rival}}$ is the
leak-free per-case-preference i.i.d.\ null (preference and accuracy from
other runs); $\phi=\agreementIndex_{\mathrm{rival}}/\agreementIndex_{\mathrm{emp}}^{(t)}$ is
the mechanical coverage on the held-out test-run population,
$\delta=1-\phi$ the preference-unexplained residual; $\rho=\agreementIndex_{\mathrm{iid}}/\agreementIndex_{\mathrm{emp}}$
is the strictly more conservative uniform null. CI column: 95\%
case-clustered percentile bootstrap ($B=10^4$), \emph{untruncated}; a CI
that reaches above $1$ (or below $0$) is reported as-is. Simulation
seed 0, $n_{\mathrm{sim}}=10^5$; Monte Carlo SE $\le4\times10^{-3}$
$\agreementIndex$ units in every cell. Per-cell sample sizes are in
Table~\ref{tab:samples}.}
\label{tab:kappa}
\begin{tabular}{llcccccc}
\toprule
Model & Prompt & $p$ & $\agreementIndex_{\mathrm{emp}}$ & $\agreementIndex_{\mathrm{rival}}$ & $\phi$ & $\phi$ CI & $\agreementIndex_{\mathrm{iid}}$ \\
\midrule
\texttt{4.1} & AIME-ZS & 0.123 & 6.26 & 4.26 & 0.687 & [0.640, 0.733] & 2.13 \\
\texttt{4.1} & GPQA-ZS & 0.474 & 4.94 & 4.51 & 0.915 & [0.879, 0.948] & 2.14 \\
\texttt{4.1-mini} & AIME-CoT & 0.375 & 5.69 & 4.15 & 0.726 & [0.703, 0.813] & 2.53 \\
\texttt{4.1-mini} & AIME-ZS & 0.260 & 6.76 & 3.96 & 0.586 & [0.554, 0.605] & 2.49 \\
\texttt{4.1-mini} & GPQA-CoT & 0.550 & 4.65 & 4.28 & 0.927 & [0.874, 0.974] & 2.54 \\
\texttt{4.1-mini} & GPQA-ZS & 0.492 & 4.54 & 3.66 & 0.806 & [0.771, 0.834] & 2.17 \\
\texttt{4.1-nano} & AIME-ZS & 0.163 & 7.09 & 5.55 & 0.781 & [0.733, 0.811] & 2.70 \\
\texttt{4.1-nano} & GPQA-ZS & 0.385 & 3.56 & 3.21 & 0.902 & [0.864, 0.938] & 1.82 \\
\bottomrule
\end{tabular}
\end{table}

\begin{figure}[H]
\centering
\includegraphics[width=0.62\textwidth]{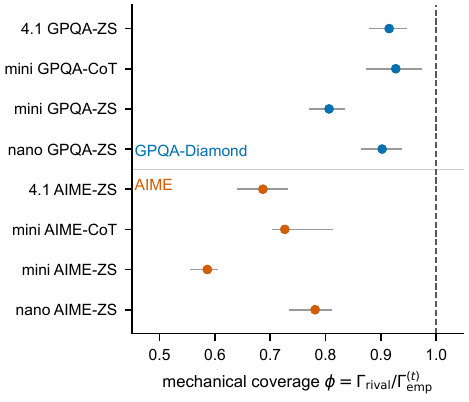}
\caption{Mechanical coverage $\phi$ by cell, with 95\% case-clustered
bootstrap CIs ($B=10^4$). Blue: GPQA-Diamond; orange: AIME (Wong colorblind-safe palette). The benchmark
difference is direction-consistent (mean difference $0.19$, $n=4$ cells per
benchmark; permutation ordering in Appendix~\ref{app:assoc}) but the boundary cells
overlap (\texttt{mini} GPQA-ZS $[0.77,0.83]$ vs.\ \texttt{nano} AIME-ZS
$[0.73,0.81]$)}
\label{fig:phi}
\end{figure}

On the gpt-4.1 family the observed pattern is a
\emph{benchmark-associated direction} in
$\phi$ (an observational ordering over $n=4$ cells per benchmark),
though the two benchmark ranges are not cleanly separated:
\begin{itemize}
\item \textbf{Multiple-choice GPQA-Diamond ($C=4$).} The mechanical
  per-case preference reproduces most of the agreement index:
  $\phi\in[0.806,0.927]$, with preference-unexplained residuals $\delta$ of
  $0.08$--$0.19$ ($0.35$--$0.88$ $\agreementIndex$ units). Here a wrong but popular
  option is largely captured by the
  per-case preference channel, so a shared-bias-dominant interpretation
  is not required to explain the observed agreement on
  multiple-choice, with the
  caveat that the origin of the preference itself (training bias or not)
  is not identified by this decomposition.
\item \textbf{Open-domain AIME ($C\approx9$--$20$).} The mechanical
  preference reproduces only $59$--$78\%$
  ($\phi\in[0.586,0.781]$), leaving a preference-unexplained residual of
  $1.54$--$2.80$ $\agreementIndex$ units ($\delta\in[0.22,0.41]$). Whereas on
  multiple-choice the set of contenders is fixed, on open-domain
  a wrong run's samples cluster on the consensus even after
  conditioning the counterfactual on the estimated per-case option preference. Appendix~\ref{app:dispersion}
  calibrates an explanatory null for this residual: a reference that grants
  each case \emph{run-level preference heterogeneity}
  (Dirichlet-multinomial, dispersion matched to the observed
  within-case cross-run spread of the plurality share) \emph{more than}
  reproduces the empirical index ($\phi_{\mathrm{dm}}=1.4$--$2.1$ on
  AIME vs.\ $0.85$--$1.01$ on GPQA). The dispersion parameter is fitted on a random \emph{held-out half} of the eligible cases and evaluated on the other half, so the parameter never touches the data it explains: the residual is more than absorbed by (hence consistent with)
  run-to-run preference variation
  on open-domain tasks, and should be read as a
  signature consistent with an upper bound, not an identified mechanism. This
  residual is the estimate most exposed to design choices: its
  sensitivity to the preference estimate (shrinkage) and to the
  option-support convention is probed in
  Appendix~\ref{app:shrink}, which shows the residual is conservative
  under shrinkage.
\end{itemize}
The GPQA/AIME split is real but modest: the four GPQA cells are the four
largest $\phi$ values, and a permutation test over the benchmark labelings
(exact enumeration in Appendix~\ref{app:assoc}) yields a small count at the
extreme of the null; we report this as a descriptive ordering, not a
significance claim (the cells are not exchangeable units: three nested model
sizes, one pipeline). The mean
difference is $0.19$ (95\% coupled bootstrap CI $[0.11,0.28]$), and the
case-clustered
CIs of the two boundary cells (\texttt{4.1-mini} GPQA-ZS
$[0.77,0.83]$ vs.\ \texttt{4.1-nano} AIME-ZS $[0.73,0.81]$) overlap. The finding is therefore described as \emph{direction-consistent and
partially overlapping}, not as a clean separation. \Cref{fig:phi} shows
all eight cells with their case-clustered CIs. The two benchmarks
also differ in difficulty (GPQA $p\in[0.38,0.55]$ vs.\ AIME
$p\in[0.12,0.38]$); within-benchmark, $\phi$ is not monotone in $p$
(on AIME the minimum $\phi=0.586$ occurs at mid-difficulty $p=0.26$,
while the lowest- and highest-$p$ cells both exceed $0.68$; on GPQA the
lowest-$p$ cell has $\phi=0.90$): the observed within-benchmark pattern is not monotonic in single-sample accuracy, so there is no simple evidence for a monotonic difficulty-only explanation. Difficulty and answer-space openness
remain confounded. An associational (no-significance) analysis of the agreement index with single-sample accuracy, answer-space size $C$, and the open/closed indicator across the eight cells, and why it is not read as a benchmark effect, is given in Appendix~\ref{app:assoc}.

Taken together, ``shared
bias'' is \emph{not} a single channel: on constrained multiple-choice
it is largely absorbed by the per-case preference, and on open-domain a
residual beyond the preference survives. The uniform null is strictly more
conservative ($\rho=\agreementIndex_{\mathrm{iid}}/\agreementIndex_{\mathrm{emp}}\in
[0.340,0.546]$ in every cell), so the observed index is never
\emph{less} mechanical than the uniform ceiling implies; the leak-free
null merely shows how much of the gap is attributable to per-case
preference rather than correlated error.

\subsection{Controlled replication: near-complete mechanical coverage across open-weights models}
\label{sec:res-tier3}

The cells of \Cref{tab:kappa} come from one model family (gpt-4.1) and
from per-run tables published by a prior study; the benchmark-associated
direction in $\phi$ may therefore be specific to that family rather than
a property of the benchmarks. To probe this, controlled sampling was run
under a fixed protocol (four runs per question, $K=32$ votes per run,
temperature $0.7$, zero-shot) on open-weights models of different
sizes and families (\texttt{qwen3.5-9b}, \texttt{qwen3.8-27b},
\texttt{gemma4-26b}, \texttt{gemma4-31b}, and a \texttt{qwen3.5-122b}
arm) across two benchmarks: the official
GPQA-Diamond set (198 questions, $C=4$) and a stratified sample of 200
AIME questions spanning 1983 to 2024. The same decomposition machinery
was applied to every cell.

The per-case preference null reproduces the empirical agreement index
with near-complete mechanical coverage in \emph{every} Tier-3 cell, on
both benchmarks and across five model sizes:
$\phi\in[1.001,1.056]$, at or slightly above $1$ in all ten cells
(\Cref{tab:tier3}). No cell falls below $1$. The four AIME cells sit at low accuracy
($p\le0.054$); the five GPQA-Diamond cells sit at
$p\in[0.40,0.49]$, overlapping the gpt-4.1 GPQA accuracy range
($[0.38,0.55]$) while their coverage is $1.00$--$1.01$, against
$\phi\in[0.81,0.93]$ for the gpt-4.1 GPQA cells: at comparable accuracy
on multiple-choice, the open-weights cells leave no
preference-unexplained residual where the frontier family leaves
$\delta=0.08$--$0.19$.

The ninth cell extends the contrast to open-domain tasks. The
\texttt{qwen3.5-122b} AIME cell operates at $p=0.222$, inside the
gpt-4.1 AIME accuracy range ($[0.12,0.38]$), and its mechanical coverage
is $\phi=1.041$ (case-clustered CI $[1.031,1.054]$), at the same
saturation as every other open-weights cell instead of inside the
gpt-4.1 AIME range ($[0.586,0.781]$). At comparable aggregate AIME accuracy,
therefore, the open-weights cell saturates where the frontier family
retains a preference-unexplained residual of $1.54$--$2.80$ $\agreementIndex$
units. The contrast of Section~\ref{sec:res-kappa} is thus neither a
benchmark property nor a pure accuracy effect. It is observed on both
benchmarks once aggregate accuracy is made comparable across families
(a five-cell GPQA bridge; a single-cell AIME bridge at \texttt{qwen3.5-122b}). Its remaining
candidate drivers are the family contrast itself (capability class,
closed versus open weights, training distribution) and the protocol
differences listed below, none of which is isolated by this design. In
the same cell majority voting mildly \emph{helps}: consensus accuracy
$0.266$ (the fraction of runs whose majority answer is correct)
against single-sample $p=0.222$, so no extreme-backfire reading
attaches to it. On
GPQA-Diamond the open-weights cells reach $p\le0.49$, so the
accuracy-matched contrast there covers nearly the whole gpt-4.1 range,
missing only its top cell (at $p\approx0.55$).

Why the open-weights values sit at, rather than below, $1$: the
leave-one-out preference $\hat q_i$ is a plug-in estimate from only
three donor runs per case (96 votes), and the simulated agreement
concentrates more under a noisy estimate than under the true preference,
so $\agreementIndex_{\mathrm{rival}}$ is biased upward and $\phi$ lands at or
above $1$. Three checks bear out this reading. First, the two-run
sensitivity reduces the donors to one and raises the grid ceiling from
$1.056$ to $1.155$, the direction the plug-in bias predicts. Second,
shrinking the preference estimate toward uniform (the
Appendix~\ref{app:shrink} method) removes the overshoot: at
$\lambda=0.5$, every spot-checked cell's coverage falls below $1$
(\texttt{qwen3.5-9b} AIME $1.056\to0.713$,
\texttt{qwen3.8-27b} AIME $1.052\to0.713$,
\texttt{qwen3.5-122b} AIME $1.041\to0.687$,
\texttt{gemma4-26b} GPQA $1.002\to0.917$;
\texttt{results/tier3\_shrink\_spot.json}). Third, and most directly, we
calibrate the bias by simulation
(\texttt{analysis/finite\_donor\_mc.py}): for each case we fit the
i.i.d.-null parameters from all observed votes and regenerate the data
from that null, then run the identical leave-one-out pipeline. With the
true per-case preference the estimator reads
$\phi=0.9997$--$1.0007$ across all ten cells (consistent with the null
within simulation noise), whereas with the leave-one-out preference it reads
$\phi$ above $1$ by a positive, cell-specific bias of $+0.001$ to
$+0.005$ (four-option GPQA) and $+0.012$ to $+0.053$ (low-accuracy AIME);
the observed $\phi$ ($1.0008$--$1.0561$) matches this biased estimate
within $0.01$ for every cell. By contrast, the \texttt{gpt-4.1} cells are calibrated with their
\emph{own} sampling structure (two runs, $K=50$, one leave-one-out donor per
case) rather than the open-weights one. Their finite-donor band still lies
above $1$ ($[1.020,1.033]$ AIME, $[1.000,1.006]$ GPQA, a positive bias of
$+0.026$ / $+0.003$ in both benchmarks) and their observed
$\phi$ ($0.586$--$0.781$ AIME, $0.806$--$0.927$ GPQA) falls far
below their \emph{own} band; their deficit is therefore not the finite-donor
plug-in effect under their own structure and remains a preference-unexplained
residual under the fixed-preference i.i.d.\ reference
(\texttt{results/finite\_donor\_mc.json}). The canonical $\lambda=1$
values are therefore read as full coverage at the unshrunk
estimate, with the overshoot above $1$ itself the quantified plug-in
artifact rather than evidence against a residual.

Two further design checks make sure these readings are not artifacts
of the sampling protocol. First, the Tier-3 protocol collects four runs
per question, so the leave-one-out preference estimate pools three runs,
whereas the gpt-4.1 cells hold out one run out of two; restricting each
Tier-3 cell to its first two runs per question (matching the published
design) leaves the coverage intact
($\phi\in[1.003,1.155]$;
\texttt{results/tier3\_kappa\_r2.json}). The two-run
experiment is a qualitative robustness check on the coverage pattern
under a reduced run count, not a precision-improving estimator. Second,
$\phi$ is $C$-invariant by construction, and every null is recomputed at
the observed vote count of its own cell, so the comparison does not
inherit the vote-count difference between the two data sources; as a
direct check on the key cell, rerunning \texttt{qwen3.5-122b} AIME
with $C$ fixed to $9$ or to $20$ leaves $\phi=1.041$ unchanged to six
decimals (\texttt{results/tier3\_csens\_122b.json}).

\begin{figure}[H]
\centering
\includegraphics[width=0.66\textwidth]{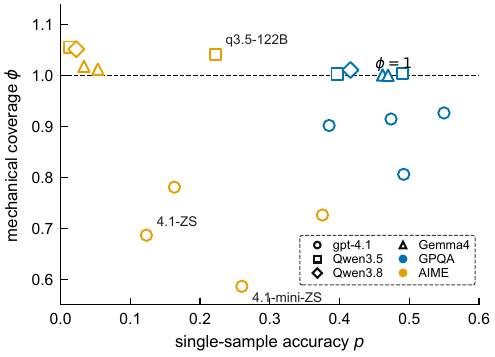}
\caption{Cross-system contrast at comparable aggregate accuracy in mechanical coverage
$\phi$ against single-sample accuracy $p$, all eighteen cells. All markers are hollow; marker shape: model family; edge
color: benchmark (blue GPQA-Diamond, orange AIME; Wong
colorblind-safe). The gpt-4.1 points come from the Ding released
pipeline ($K=50$, temperature unrecorded) and the open-weights points
from the controlled protocol of \Cref{sec:res-tier3} ($K=32$,
temperature $0.7$); the protocol difference coincides with the family
split here, so it is disclosed textually instead of being visually
encoded. No
regression or connectivity line is drawn: the comparison is a
cross-system contrast at comparable aggregate accuracy, not an estimated functional
relationship. Values at or slightly above the dashed line ($\phi=1$) mean the
fixed-preference counterfactual reproduces the observed wrong-consensus
agreement to within simulation and plug-in estimation uncertainty,
leaving no evidence for a positive preference-unexplained residual;
 values substantially below it leave one. The overshoot above $1$ is
consistent with a finite-donor plug-in bias in the preference estimate
(\Cref{sec:res-tier3}), so the open-weights cluster is read as
full mechanical coverage; 95\%
case-clustered CIs are reported in \Cref{tab:kappa} and
\Cref{tab:tier3}; they are not drawn here.
Text labels mark the \texttt{qwen3.5-122b} AIME point, the one with AIME accuracy inside the gpt-4.1 range
($p=0.222$), the minimum-$\phi$ gpt-4.1 AIME cell (\texttt{4.1-mini}
zero-shot, $\phi=0.586$), and the \texttt{4.1} AIME cell. The ten
open-weights cells cluster at $\phi\approx1$ across
$p\in[0.013,0.49]$, while the eight gpt-4.1 cells sit below $1$
($\phi\in[0.586,0.927]$)}
\label{fig:regime}
\end{figure}

The reading is therefore a cross-system
contrast at comparable aggregate accuracy, not a benchmark or accuracy
regime statement. The
wrong-consensus agreement of the open-weights models saturates
on both benchmarks across all observed cells
($\phi\approx1$); the frontier family, in contrast, retains a
preference-unexplained residual in every measured cell. The contrast
replicates on both benchmarks when accuracy is matched across families.
Capability class, closed versus open weights, training distribution, and
protocol differences (vote count, temperature) are confounded
with the family split by design: every gpt-4.1 cell is drawn from the
runner-based public pipeline and every open-weights cell from the fixed
controlled protocol, so family and protocol cannot be separated on the
present data. No
mechanism is identified.
\Cref{fig:regime} shows all eighteen cells together.

\begin{table}[H]
\centering
\caption{Tier-3 controlled replication. Fixed protocol: four runs per
question, $K=32$ votes per run, temperature $0.7$, zero-shot; official
GPQA-Diamond (198 questions) and a stratified AIME sample (200 questions,
1983--2024). Columns as in \Cref{tab:kappa}; $\phi$ with case-clustered
95\% CI ($B=10^4$; both interval constructions in
Appendix~\ref{app:ci}); simulation seed 0, $n_{\mathrm{sim}}=10^5$. In
every cell the case-clustered interval brackets the point estimate.}
\label{tab:tier3}
\begin{tabular}{llcccccc}
\toprule
Model & Prompt & $p$ & $\agreementIndex_{\mathrm{emp}}$ & $\agreementIndex_{\mathrm{rival}}$ & $\phi$ & $\phi$ CI & $\agreementIndex_{\mathrm{iid}}$ \\
\midrule
\texttt{qwen3.5-9b-ctx4k} & GPQA-ZS & 0.397 & 3.35 & 3.36 & 1.003 & [1.000, 1.007] & 1.92 \\
\texttt{gemma4-26b} & GPQA-ZS & 0.462 & 4.82 & 4.83 & 1.002 & [1.000, 1.004] & 2.28 \\
\texttt{qwen3.8-27b} & GPQA-ZS & 0.416 & 2.98 & 3.01 & 1.011 & [1.006, 1.017] & 1.92 \\
\texttt{gemma4-31b} & GPQA-ZS & 0.469 & 5.26 & 5.26 & 1.001 & [1.000, 1.002] & 2.35 \\
\texttt{qwen3.5-9b} & AIME-ZS & 0.013 & 3.96 & 4.19 & 1.056 & [1.045, 1.069] & 2.67 \\
\texttt{gemma4-26b} & AIME-ZS & 0.033 & 4.68 & 4.77 & 1.018 & [1.014, 1.024] & 2.00 \\
\texttt{qwen3.8-27b} & AIME-ZS & 0.023 & 3.97 & 4.18 & 1.052 & [1.042, 1.064] & 2.47 \\
\texttt{gemma4-31b} & AIME-ZS & 0.054 & 4.10 & 4.16 & 1.013 & [1.009, 1.018] & 1.86 \\
\texttt{qwen3.5-122b} & AIME-ZS & 0.222 & 4.65 & 4.84 & 1.041 & [1.031, 1.054] & 2.89 \\
\texttt{qwen3.5-122b} & GPQA-ZS & 0.490 & 3.92 & 3.93 & 1.004 & [1.000, 1.008] & 2.27 \\
\bottomrule
\end{tabular}
\end{table}

\subsection{Complementary diagnostics (backfire, fragility, Jensen gap)}
\label{sec:res-backfire}

Three complementary signatures of the wrong-consensus regime are reported
in Appendix~\ref{app:backfire}--\ref{app:jensen}: (i)~the backfire
phenomenon of \cite{bahuguna2026when} is reproduced in difficulty-binned
form (voting gap negative on the hardest bins, CI excluding zero for two
cells); (ii)~the champion flip rate is substantial ($0.40$ GPQA, $0.82$
AIME for \texttt{gpt-4.1-mini}) and high agreement only weakly stabilizes
the winner; (iii)~soft agreement adds no AUROC over hard majority
($0.769$ vs.\ $0.770$ pooled), and the Jensen gap is roughly twice as
large on wrong runs ($0.055$) as correct ones ($0.027$). These are
reported as understanding diagnostics, not as new voting methods.

\subsection{Pilot-scale stability of the diagnostics}
\label{sec:res-pilot}

A $25$-case pilot already returns the coverage reading and preserves the
cell ordering (Tier-3 $\phi$ within $0.010$ of the full value,
$r_s=0.94$; gpt-4.1 $\agreementIndex$ $r_s=0.976$, $\phi$ $r_s=0.786$); the full
protocol, tables, and descriptive $p$-values are in
Appendix~\ref{app:pilot}. Its intended use is a pre-decision diagnostic,
not per-question prediction.

\section{Limitations}
\label{sec:limitations}

The principal limitation is the i.i.d.\ assumption (Section~\ref{sec:iid}):
temperature draws are correlated. The effect of such dependence on the
conditional wrong-consensus index is not identified by this design, so
both counterfactuals are treated as benchmark
references rather than certified bounds, and the residual $\delta$ is
read as a residual relative to the i.i.d.\ reference, not as a bound on
an underlying correlation-free residual; it is
an observational signature, not a point-identified mechanism. External
validity: the main text is built on the GPT-4.1 family; the controlled
Tier-3 replication (Section~\ref{sec:res-tier3}) extends the rival
decomposition to open-weights models across both benchmarks and finds
full mechanical saturation in all ten cells, including two with accuracy
inside the gpt-4.1 range (AIME $p=0.222$, GPQA $p=0.490$), so the residual is neither a
benchmark property nor a pure accuracy effect but a cross-system
contrast. The candidate drivers behind it (capability class, closed versus open weights, training distribution, and protocol differences) are confounded with the family split by design (Section~\ref{sec:res-tier3}), so a protocol reading and a family reading cannot be separated without a controlled frontier run. The
reading stays observational and nothing is identified. One asymmetry bounds the contrast: on
GPQA-Diamond no open-weights cell reaches above $p=0.49$ (the contrast is
tested over nearly the whole gpt-4.1 range there, missing only its
extreme top), and the
frontier side rests on one family (gpt-4.1, three sizes) from a
pipeline whose temperature settings are not recorded and whose
within-case runs come from many independent runners with their own
prompts, so runner heterogeneity is a candidate channel for the
frontier residual that the controlled Tier-3 protocol excludes by
construction. The smaller
Qwen3.5-9B arm of Appendix~\ref{app:qwen} ($n=70$ on MMLU-Pro, no CIs)
remains a direction-consistent uniform share only. The difficulty-binned backfire is a small-sample analysis for some
cells; bootstrap CIs are reported and underpowered subsets are not
over-interpreted. No fix is proposed; the boundary is that this
paper is an understanding contribution.

\section{Discussion}
\label{sec:discussion}

Nonetheless, the pattern is consistent and practically relevant across the robustness checks. The decomposition is
\emph{benchmark-associated in direction} on the gpt-4.1 family: on multiple-choice GPQA-Diamond the
mechanical per-case answer preference accounts for $\approx81$--$93\%$
of the agreement index; on open-domain AIME the mechanical preference
accounts for
only $59$--$78\%$ and a preference-unexplained residual of $1.54$--$2.80$ $\agreementIndex$
units survives (the four GPQA cells are the four largest $\phi$ values;
permutation ordering in Appendix~\ref{app:assoc}, $n=4$ per benchmark;
Section~\ref{sec:res-kappa}).
Benchmark differences, however, are insufficient to explain the
variation: the controlled replication of Section~\ref{sec:res-tier3}
shows full mechanical coverage in all ten open-weights cells
across both benchmarks, including two with accuracy inside the gpt-4.1 range
(AIME and GPQA),
so the residual is neither a benchmark property nor a pure accuracy
effect but a cross-system contrast, with capability class and training
regime left inseparable.
This reframes self-consistency confidence: on
constrained multiple-choice, a popular-but-wrong answer is captured by
the per-case preference channel (the whole cohort is attracted to it),
though the origin of that preference is not identified; on open-domain,
high agreement
carries a substantial preference-unexplained component, which
Appendix~\ref{app:dispersion} shows is more than absorbed by a calibrated run-level
preference-heterogeneity null, consistent with run-to-run preference
heterogeneity and other dependence beyond a fixed per-case marginal
(\cite{mccoy2024} describes pretraining-shaped error patterns in the
same spirit); the origin of the residual remains unidentified. Either
way,
high agreement is graded evidence of correctness, not certification: the
samples of wrong runs are themselves attracted to the consensus, so
agreement saturates well below certainty (ceiling $0.42$--$0.83$;
Appendix~\ref{app:backfire}).

What this design does identify is limited to the following four points. (i)~The correlation-free reference reproduces over $80\%$ of the GPQA
index, so a shared-bias-dominant interpretation is not required to
explain the observed agreement on those cells
(Section~\ref{sec:iid}).
(ii)~The wrong-run marginal is benchmark-associated in direction, an
observational ordering at $n=4$ cells per benchmark
(Section~\ref{sec:res-kappa}). (iii)~The within-case cross-run
dispersion of the plurality share is $2$--$3\times$ larger on AIME than
GPQA, a dispersion signature that does not depend on the dispersion
null's fit (Appendix~\ref{app:dispersion}). (iv)~High agreement is
graded evidence of correctness, never certification: the empirical
high-agreement accuracy is $0.42$--$0.83$ with $1.2$--$3.6\times$ lift
(Appendix~\ref{app:backfire}), and the champion flip rate
stays substantial even at high agreement ($0.17$ GPQA at $\alpha\ge0.98$;
$0.78$ AIME at $\alpha\ge0.84$)
(Appendix~\ref{app:fragility}).

Falsifiability. The headline $\phi$ (or, more conservatively, the
uniform $\rho$) is refutable in principle: if the per-case preference
reference were sufficient, then $\phi$ would concentrate near $1$ and the shared
residual $\agreementIndex_{\mathrm{emp}}^{(t)}-\agreementIndex_{\mathrm{rival}}$, and hence
$\delta=1-\phi$, would not survive. The observed
$\phi\in[0.586,0.927]$ on gpt-4.1 is heterogeneous in direction: on GPQA it
approaches
$1$ (the preference reference largely suffices, so a \emph{correlated-error}
reading of multiple-choice is unsupported), while on AIME
$\phi\in[0.586,0.781]$
leaves a residual that survives; this rejects the sufficiency of the
fixed-preference i.i.d.\ counterfactual for the observed open-domain
agreement. The Tier-3 cells sharpen
the refutation logic: for open-weights models the reference suffices
($\phi\approx1$ across all ten cells, including one with AIME accuracy
inside the gpt-4.1 range), so the surviving residual of the frontier
family is located in the cross-system contrast rather than being a benchmark
constant or a pure accuracy effect. The paper is explicit about what this does \emph{not} rule out: positive within-case sampling
correlation is a third channel that the i.i.d.\ reference does not
incorporate, and the design cannot separate it from $\delta$. The claim
is accordingly a \emph{channel} decomposition of the wrong-run marginal
(per-case preference vs.\ residual), not a decomposition of error
\emph{sources}. Thus the claim that GPT-4.1 self-consistency
saturation is \emph{largely} mechanical on multiple-choice but carries a
\emph{robust} residual on open-domain is a falsifiable
empirical statement about the wrong-run marginal, not a tautology about
correlated draws.

\section{Conclusion}

Wrong-consensus agreement in LLM self-consistency is decomposed in this paper into a
mechanical plurality effect and a preference-unexplained residual,
measured by the mechanical coverage $\phi$. On constrained
multiple-choice GPQA-Diamond, a leak-free per-case preference reference
reproduces over $80\%$ of the agreement index, a benchmark-mechanical
saturation; on open-domain AIME a preference-unexplained residual
survives ($\phi$ as low as $0.586$), which a calibrated run-heterogeneity
null more than absorbs (Appendix~\ref{app:dispersion}), consistent with
run-to-run preference heterogeneity and shared dependence beyond a fixed
per-case marginal; the underlying source is not identified by this design. Across model families,
the controlled replication shows the contrast in its cleanest form:
near-complete mechanical coverage in all ten open-weights cells,
including one with AIME accuracy inside the gpt-4.1 range, while the frontier
family retains its residual at every measured cell
(Section~\ref{sec:res-tier3}). The result reframes
self-consistency confidence as graded evidence of correctness, never
certification. These conclusions are bounded by the i.i.d.\ assumption;
the rival null is extended across model families by the Tier-3
replication (Section~\ref{sec:res-tier3}), but the frontier side of the
contrast rests on one family (gpt-4.1) and its AIME residual is tested
against a single matched-accuracy open-weights cell
(Qwen3.5-122B); future work should add independent frontier families,
separate within-case sampling correlation from $\delta$, and turn
the channel decomposition into explicit uncertainty estimates.

\paragraph{Statements.}
Data and code availability: the gpt-4.1 per-run data are a public
release of Ding~\cite{ding2026auditing}; the Tier-3 per-sample data were
collected for this paper under the protocol of
Section~\ref{sec:res-tier3} and are committed as raw JSONL evidence.
All analysis scripts and evidence
files (JSON) that produced every number in this paper are committed in full (scripts and data tracked in the project's version control; an anonymized copy will be made available through the venue's anonymous-repository upload, e.g.\ the review system's anonymized-upload field, so reviewers can access it directly rather than on request),
and the analysis can be regenerated from the raw files
by the commands in each script's header. Funding: supported in part
by the Yuelushan Laboratory Breeding Program (Grant YLS-2026-ZY01002), in
part by the National Natural Science Foundation of China (Grant
62372064), and in part by the Meizhou Tobacco Science Research Project
(Grant 202404). The funders had no role in study design, data collection
and analysis, decision to publish, or preparation of the manuscript.
Competing interests: none declared. Ethics: the gpt-4.1 analysis
re-analyzes existing model outputs; the Tier-3 replication
(Section~\ref{sec:res-tier3}) performs new inference with open-weights
models on public benchmarks (GPQA-Diamond, AIME). No human subjects are
involved.

\section*{Acknowledgments}
This work was supported in part by the Yuelushan Laboratory Breeding
Program under Grant YLS-2026-ZY01002, in part by the National Natural
Science Foundation of China under Grant 62372064, and in part by the
Meizhou Tobacco Science Research Project under Grant 202404.

\bibliographystyle{elsarticle-num}
\bibliography{refs}

@inproceedings{wang2022selfconsistency,
  title={Self-Consistency Improves Chain of Thought Reasoning in Language Models},
  author={Wang, Xuezhi and Wei, Jason and Schuurmans, Dale and Le, Quoc and Chi, Ed and Narang, Sharan and Chowdhery, Aakanksha and Zhou, Denny},
  booktitle={International Conference on Learning Representations (ICLR)},
  year={2023},
  note={arXiv:2203.11171}
}

@misc{ding2026auditing,
  title={When {LLM}s Agree, Are They Right? {Auditing Self-Consistency and Cross-Model Agreement as Confidence Signals}},
  author={Ding, Kaihua},
  year={2026},
  eprint={2607.08065},
  archivePrefix={arXiv},
  primaryClass={cs.AI},
}

@misc{bahuguna2026when,
  title={When Self-Consistency Backfires: Majority Vote Hurts the Majority of Hard Science Problems for Small {LLM}s},
  author={Bahuguna, Utkarsh},
  year={2026},
  eprint={2608.11403},
  archivePrefix={arXiv},
  primaryClass={cs.AI},
  note={v2 of 2026-08-15}
}

@misc{anchorscore2026,
  title={{AnchorScore}: A {CLIP}-Based Diagnostic of {MLLM} Annotation Difficulty},
  author={Ma, Yan and Zhang, Lizhuo},
  year={2026},
  eprint={2608.16690},
  archivePrefix={arXiv},
  primaryClass={cs.CV},
}

@article{cohen1960,
  title={A Coefficient of Agreement for Nominal Scales},
  author={Cohen, Jacob},
  journal={Educational and Psychological Measurement},
  volume={20},
  number={1},
  pages={37--46},
  year={1960},
  doi={10.1177/001316446002000104}
}

@article{scott1955,
  title={Reliability of Content Analysis: The Case of Nominal Scale Coding},
  author={Scott, William A.},
  journal={Public Opinion Quarterly},
  volume={19},
  number={3},
  pages={321--325},
  year={1955},
  doi={10.1086/266577}
}

@article{fleiss1971,
  title={Measuring Nominal Scale Agreement Among Many Raters},
  author={Fleiss, Joseph L.},
  journal={Psychological Bulletin},
  volume={76},
  number={5},
  pages={378--382},
  year={1971},
  doi={10.1037/h0031619}
}

@article{krippendorff1970,
  title={Bivariate Agreement Coefficients for Reliability of Data},
  author={Krippendorff, Klaus},
  journal={Sociological Methodology},
  volume={2},
  pages={139--150},
  year={1970},
  doi={10.2307/270787}
}

@article{kuncheva2003,
  title={Measures of Diversity in Classifier Ensembles and Their
  Relationship with the Ensemble Accuracy},
  author={Kuncheva, Ludmila I. and Whitaker, Christopher J.},
  journal={Machine Learning},
  volume={51},
  number={2},
  pages={181--207},
  year={2003},
  doi={10.1023/A:1022859003006}
}

@inproceedings{kuhn2023,
  title={Semantic Uncertainty: Linguistic Invariances for Uncertainty Estimation in Natural Language Generation},
  author={Kuhn, Lorenz and Gal, Yarin and Farquhar, Sebastian},
  booktitle={International Conference on Learning Representations (ICLR)},
  year={2023}
}

@article{farquhar2024,
  title={Detecting hallucinations in large language models using semantic entropy},
  author={Farquhar, Sebastian and Kossen, Jannik and Kuhn, Lorenz and Gal, Yarin},
  journal={Nature},
  volume={630},
  pages={625--630},
  year={2024},
  doi={10.1038/s41586-024-07421-0}
}

@inproceedings{manakul2023,
  title={{SelfCheckGPT}: Zero-Resource Black-Box Hallucination Detection for Generative Large Language Models},
  author={Manakul, Potsawee and Liusie, Adian and Gales, Mark J. F.},
  booktitle={Proceedings of the 2023 Conference on Empirical Methods in Natural Language Processing (EMNLP)},
  pages={9004--9017},
  year={2023},
  doi={10.18653/v1/2023.emnlp-main.557}
}

@article{mccoy2024,
  title={Embers of autoregression show how large language models are shaped by the problem they are trained to solve},
  author={McCoy, R. Thomas and Yao, Shunyu and Friedman, Dan and Hardy, Mathew D. and Griffiths, Thomas L.},
  journal={Proceedings of the National Academy of Sciences},
  volume={121},
  number={41},
  pages={e2322420121},
  year={2024},
  doi={10.1073/pnas.2322420121}
}

@inproceedings{kohavi1996,
  title={Bias plus Variance Decomposition for Zero-One Loss Functions},
  author={Kohavi, Ron and Wolpert, David H.},
  booktitle={International Conference on Machine Learning (ICML)},
  pages={275--283},
  year={1996}
}

@inproceedings{krogh1995,
  title={Neural Network Ensembles, Cross Validation, and Active Learning},
  author={Krogh, Anders and Vedelsby, Jesper},
  booktitle={Advances in Neural Information Processing Systems (NIPS)},
  pages={231--238},
  year={1994}
}

@inproceedings{ueda1996,
  title={Generalization Error of Ensemble Estimators},
  author={Ueda, Naonori and Nakano, Ryohei},
  booktitle={IEEE International Conference on Neural Networks (ICNN)},
  pages={90--95},
  year={1996},
  doi={10.1109/ICNN.1996.548872}
}

@misc{hamidieh2026,
  title={Complementing Self-Consistency with Cross-Model Disagreement for Uncertainty Quantification},
  author={Hamidieh, Kimia and Thost, Veronika and Gerych, Walter and Yurochkin, Mikhail and Ghassemi, Marzyeh},
  year={2026},
  eprint={2604.17112},
  archivePrefix={arXiv},
  primaryClass={cs.AI},
}

@misc{fadnavis2026,
  title={Beyond Consensus: Trace-Level Synthesis in Mixture of Agents},
  author={Fadnavis, Shreyas and Kanakaraj, Praitayini and Wyss, Felix},
  year={2026},
  eprint={2605.29116},
  archivePrefix={arXiv},
  primaryClass={cs.AI},
}

@misc{kim2026,
  title={Are Diversity Metrics Measuring Diversity? {A Capability-Controlled Audit of Majority-Vote Gain in {LLM} Ensembles}},
  author={Kim, Donghwan},
  year={2026},
  eprint={2607.20768},
  archivePrefix={arXiv},
  primaryClass={cs.CL},
}

@inproceedings{xiao2025,
  title={The Consistency Hypothesis in Uncertainty Quantification for Large Language Models},
  author={Xiao, Quan and Bhattacharjya, Debarun and Ganesan, Balaji and Marinescu, Radu and Mirylenka, Katsiaryna and Pham, Nhan H. and Glass, Michael and Lee, Junkyu},
  booktitle={Conference on Uncertainty in Artificial Intelligence (UAI)},
  pages={4636--4651},
  year={2025},
  eprint={2506.21849},
  archivePrefix={arXiv},
  primaryClass={cs.CL},
}

@misc{nguyen2025,
  title={Beyond Semantic Entropy: Boosting {LLM} Uncertainty Quantification with Pairwise Semantic Similarity},
  author={Nguyen, Dang and Payani, Ali and Mirzasoleiman, Baharan},
  year={2025},
  eprint={2506.00245},
  archivePrefix={arXiv},
  primaryClass={cs.LG},
}

@misc{arzhantsev2026,
  title={Self-Consistency via Marginal Sharpening},
  author={Arzhantsev, Aleksei and Sakhi, Otmane and Chopin, Nicolas},
  year={2026},
  eprint={2605.28142},
  archivePrefix={arXiv},
  primaryClass={cs.LG},
}

@inproceedings{tan2025,
  title={Too Consistent to Detect: A Study of Self-Consistent Errors in {LLM}s},
  author={Tan, Hexiang and Sun, Fei and Liu, Sha and Su, Du and Cao, Qi and Chen, Xin and Wang, Jingang and Cai, Xunliang and Wang, Yuanzhuo and Shen, Huawei and Cheng, Xueqi},
  booktitle={Empirical Methods in Natural Language Processing (EMNLP)},
  pages={4755--4765},
  year={2025},
  doi={10.18653/v1/2025.emnlp-main.238},
  eprint={2505.17656},
  archivePrefix={arXiv},
  primaryClass={cs.CL},
}

@inproceedings{chen2024,
  title={Are More {LLM} Calls All You Need? {Towards the Scaling Properties of Compound {AI} Systems}},
  author={Chen, Lingjiao and Davis, Jared Quincy and Hanin, Boris and Bailis, Peter and Stoica, Ion and Zaharia, Matei and Zou, James},
  booktitle={Advances in Neural Information Processing Systems (NeurIPS)},
  volume={37},
  pages={45767--45790},
  year={2024},
  eprint={2403.02419},
  archivePrefix={arXiv},
  primaryClass={cs.LG},
  doi={10.52202/079017-1455},
}

@misc{corderoencinar2025certified,
  title={Certified {Self-Consistency}: {Statistical} {Guarantees} and {Test-Time} {Training} for {Reliable} {Reasoning} in {LLMs}},
  author={Cordero-Encinar, Paula and Duncan, Andrew B.},
  year={2025},
  eprint={2510.17472},
  archivePrefix={arXiv},
  primaryClass={cs.LG},
}

@inproceedings{kim2025correlated,
  title={Correlated {Errors} in {Large} {Language} {Models}},
  author={Kim, Elliot and Garg, Avi and Peng, Kenny and Garg, Nikhil},
  booktitle={International Conference on Machine Learning (ICML)},
  year={2025},
  eprint={2506.07962},
  archivePrefix={arXiv},
  primaryClass={cs.LG},
}

@inproceedings{ali2026stochastic,
  title={Stochastic {Sampling} is {Epistemically} {Shallow}: {The} {Dimensionality} {Gap} {Between} {Temperature} {Variation} and {Model} {Diversity} in {LLMs}},
  author={Ali, Izhar},
  booktitle={Explainable, Interpretable, and Transparent ML Workshop (EIML) at ICML},
  year={2026},
  eprint={2607.20464},
  archivePrefix={arXiv},
  primaryClass={cs.LG},
}

\clearpage
\appendix

\section{The pooled-$p$ control fails}
\label{app:pooled}

This appendix deliberately reports a counterfactual that \emph{fails}, because its
failure is what justifies difficulty-matching in both i.i.d.\ nulls of
Section~\ref{sec:iid}. \Cref{tab:pooled} gives, for each
cell, the empirical share of runs whose plurality label is wrong
($\mathrm{wc}_{\mathrm{emp}}$), the share predicted by the
difficulty-matched reference ($\mathrm{wc}_{\mathrm{perq}}$), and the
share predicted by a pooled-$p$ reference
($\mathrm{wc}_{\mathrm{pooled}}$).

\begin{table}[H]
\centering
\caption{Pooled-$p$ is a failed control. $\mathrm{wc}$ is the share of runs whose plurality label is wrong: empirical ($\mathrm{wc}_{\mathrm{emp}}$), predicted by the difficulty-matched i.i.d.\ reference ($\mathrm{wc}_{\mathrm{perq}}$), and predicted by a pooled-$p$ reference ($\mathrm{wc}_{\mathrm{pooled}}$). A difficulty-matched i.i.d.\
reference reproduces the empirical consensus--wrong share; a pooled-$p$
reference collapses it (by $0.8$ to $3.2$ orders of magnitude on GPQA).}
\label{tab:pooled}
\begin{tabular}{llccc}
\toprule
Model & Prompt & $\mathrm{wc}_{\mathrm{emp}}$ &
$\mathrm{wc}_{\mathrm{perq}}$ & $\mathrm{wc}_{\mathrm{pooled}}$ \\
\midrule
\texttt{4.1} & AIME-ZS & 0.833 & 0.710 & 0.568 \\
\texttt{4.1} & GPQA-ZS & 0.520 & 0.462 & 0.006 \\
\texttt{4.1-mini} & AIME-CoT & 0.558 & 0.398 & 0.001 \\
\texttt{4.1-mini} & AIME-ZS & 0.694 & 0.363 & 0.020 \\
\texttt{4.1-mini} & GPQA-CoT & 0.419 & 0.292 & 0.000 \\
\texttt{4.1-mini} & GPQA-ZS & 0.484 & 0.344 & 0.003 \\
\texttt{4.1-nano} & AIME-ZS & 0.771 & 0.640 & 0.130 \\
\texttt{4.1-nano} & GPQA-ZS & 0.598 & 0.464 & 0.088 \\
\bottomrule
\end{tabular}
\end{table}

On three of four GPQA cells, pooling over difficulty pushes the predicted
wrong-consensus share to $\le0.6\%$ (and to $8.8\%$ on
\texttt{4.1-nano} GPQA-ZS), whereas the observed share is
$42$--$60\%$ and the difficulty-matched reference recovers
$29$--$46\%$. The pooled-$p$ reference is thus a poor counterfactual for
LLM data: it attributes essentially all correctness to a single global
accuracy, treating every case as if it were of average difficulty, and thereby
understates by $1.5\times$ to over $1000\times$ ($0.2$ to $3.2$ orders of magnitude) across cells how often a \emph{wrong} plurality
forms. Note also that even the difficulty-matched reference falls short
of the empirical wrong-consensus share by
$0.06$--$0.33$ in every cell (largest on AIME, e.g.\ $0.363$ vs.\ $0.694$
for \texttt{4.1-mini} AIME-ZS): this shortfall is consistent with the additional cross-run
structure captured by the $\agreementIndex$ decomposition as $\delta>0$ on open-domain
cells, so Appendix A and Table~\ref{tab:kappa} are mutually consistent.
Any agreement mechanism estimated against a pooled-$p$ reference would be
structurally misled; this is why a difficulty-matched reference is the
informative baseline for the $\agreementIndex$ decomposition on LLMs.

\section{Pilot-sample stability of the diagnostics}
\label{app:pilot}

The agreement diagnostics ($\agreementIndex_{\mathrm{emp}}$, $\phi$) are read
\emph{before} a decision: before committing a vote budget to a new
model or benchmark, before interpreting observed agreement as
confidence. A natural operational question follows: how many pilot cases
are needed before these quantities are stable, and does a cheap pilot
order cells the same way the full dataset does? To answer it, cases (not
runs) are subsampled without replacement at sizes
$10,25,50,100,200$ and full, with deterministic seeds; $\agreementIndex_{\mathrm{emp}}$
is exact on every subsample (no simulation), and $\phi$ is computed at a
documented reduced precision ($n_{\mathrm{sim}}=2\times10^4$, bootstrap
$2\times10^3$; the canonical values of this paper remain the
$n_{\mathrm{sim}}=10^5$ ones). Pilot-vs-full agreement is scored by
Spearman rank correlation over the eight gpt-4.1 cells.

The pilot analysis covers the six Tier-3 cells completed at the time it was run (the four-model grid minus the 31B cells and the 27B AIME cell, plus the 122B cell); the later cells follow the same saturation pattern at full data, and we omit their pilot entries only for scheduling reasons. \Cref{tab:pilot} reports the pilot means and the full-data values. Two
readings hold. First, a $25$-case pilot orders cells by
$\agreementIndex_{\mathrm{emp}}$ almost exactly as the full data does
($r_s=0.976$, descriptive); the pilot mean sits within $3\%$ of the full
value in every cell, and the cross-seed standard deviation of the
$25$-case estimate is $0.26$--$0.76$ $\agreementIndex$ units, against
$0.55$--$1.48$ at $10$ cases. Twenty-five cases is therefore the point
at which the empirical index stabilizes for the coverage reading. Second, $\phi$
stabilizes more slowly but still usefully: the $25$-case pilot mean
deviates from the full-data $\phi$ by $0.012$--$0.078$ (most cells
within $0.03$), and the cell ordering is preserved ($r_s=0.786$,
descriptive); the additional noise relative to $\agreementIndex_{\mathrm{emp}}$
inherits the leave-one-out preference estimation and the reduced
simulation budget. The reading of a $25$-case pilot is already the
coverage reading the paper relies on: full mechanical coverage
($\phi\approx1$) versus a residual ($\phi<1$) is visible from the
pilot alone, without waiting for a full cell.

The intended use is explicitly bounded. This is a
\emph{pre-decision diagnostic} (does this new cell sit at
$\phi\approx1$, or is a residual present that deserves the dispersion
apparatus of Appendix~\ref{app:dispersion}?), not a per-question
prediction, and the eight-cell ordering statistics are descriptive over
a single model family. The same protocol applied to the six completed
Tier-3 cells of Section~\ref{sec:res-tier3} returns the same verdict
(bottom rows of \Cref{tab:pilot}): on the six Tier-3 cells the
$25$-case pilot mean of $\phi$
sits within $0.010$ of the full-data value in every cell, and the
coverage reading (saturated, $\phi\approx1$) is returned by the
pilot in every case, which is the use the pilot is meant for.
The cell ordering by $\agreementIndex_{\mathrm{emp}}$ is reproduced
($r_s=0.94$, $n=6$, descriptive), and the $\phi$ ordering is
preserved ($r_s=0.94$, descriptive). The distinction the pilot resolves
is thus the same one the paper draws between the open-weights cells
($\phi\approx1$) and the gpt-4.1 cells ($\phi<1$).

\begin{table}[H]
\centering
\caption{Pilot-sample stability. $\agreementIndex_{\mathrm{emp}}$ is exact on every
case subsample; $\phi$ uses the documented reduced precision
($n_{\mathrm{sim}}=2\times10^4$). Pilot columns are means over $20$
seeds ($\agreementIndex$) and $5$ seeds ($\phi$) at $25$ cases. Spearman
pilot(25)-vs-full, descriptive (no significance claim): over the eight
gpt-4.1 cells, $\agreementIndex$ $r_s=0.976$, $\phi$ $r_s=0.786$; over the six
Tier-3 cells covered by the pilot analysis, $\agreementIndex$ $r_s=0.94$, $\phi$
$r_s=0.94$. In the six Tier-3 rows, $\phi$(full) is the reduced-precision value
($n_{\mathrm{sim}}=2\times10^4$), not the canonical
$n_{\mathrm{sim}}=10^5$ value of \Cref{tab:tier3}.}
\label{tab:pilot}
\begin{tabular}{llcccc}
\toprule
Model & Prompt & $\agreementIndex_{\mathrm{emp}}(25)$ & $\agreementIndex_{\mathrm{emp}}$(full) & $\phi(25)$ & $\phi$(full) \\
\midrule
\texttt{4.1} & AIME-ZS & 6.27 & 6.26 & 0.711 & 0.688 \\
\texttt{4.1} & GPQA-ZS & 5.02 & 4.94 & 0.901 & 0.920 \\
\texttt{4.1-mini} & AIME-CoT & 5.84 & 5.69 & 0.748 & 0.729 \\
\texttt{4.1-mini} & AIME-ZS & 6.82 & 6.76 & 0.610 & 0.584 \\
\texttt{4.1-mini} & GPQA-CoT & 4.59 & 4.65 & 0.849 & 0.927 \\
\texttt{4.1-mini} & GPQA-ZS & 4.62 & 4.54 & 0.781 & 0.808 \\
\texttt{4.1-nano} & AIME-ZS & 7.01 & 7.09 & 0.854 & 0.779 \\
\texttt{4.1-nano} & GPQA-ZS & 3.64 & 3.56 & 0.917 & 0.905 \\
\midrule
\texttt{qwen3.5-9b-ctx4k} & GPQA-ZS & 3.47 & 3.35 & 1.002 & 1.003 \\
\texttt{gemma4-26b} & GPQA-ZS & 5.03 & 4.82 & 1.003 & 1.002 \\
\texttt{qwen3.8-27b} & GPQA-ZS & 2.98 & 2.98 & 1.011 & 1.011 \\
\texttt{qwen3.5-9b} & AIME-ZS & 3.74 & 3.96 & 1.065 & 1.056 \\
\texttt{gemma4-26b} & AIME-ZS & 4.64 & 4.68 & 1.020 & 1.018 \\
\texttt{qwen3.5-122b} & AIME-ZS & 4.68 & 4.65 & 1.036 & 1.041 \\
\bottomrule
\end{tabular}
\end{table}

\section{Descriptive association with cell-level descriptors}
\label{app:assoc}
As a continuous complement to the binary benchmark comparison, across the eight gpt-4.1 cells the agreement index $\agreementIndex$ (and $\phi$) is lower on hard, open-domain, high-cardinality cells: descriptive Spearman $r_s=0.81$ vs.\ single-sample accuracy, $-0.76$ vs.\ answer-space size $C$, and $-0.87$ vs.\ the open/closed indicator ($n=8$, no significance claim). The three candidate drivers are themselves strongly confounded with each other, so the data cannot separate difficulty from answer-space openness. We therefore report this as an associational, direction-consistent pattern, not an identified benchmark effect. Additionally, an exact permutation enumeration over the $\binom{8}{4}=70$ benchmark labelings gives $p=2/70\approx0.029$ (the minimum count attainable at $n=4$ per benchmark); because the eight cells are not exchangeable units (three nested model sizes, one shared pipeline), this count is a descriptive ordering statistic and we make no significance claim from it.

\section{Interval construction for the Tier-3 cells}
\label{app:ci}

\Cref{tab:tier3} reports the case-clustered percentile bootstrap
(resampling cases with replacement, coupled numerator/denominator;
Convention 5). \Cref{tab:ci} reports that construction alongside the
run-level coupled bootstrap (resampling test runs with replacement,
coupled) for all ten cells, as a transparency disclosure: a ratio of
case-heterogeneous quantities can be sensitive to the resampling unit,
and the comparison shows the coverage reading is insensitive to that
choice (in every cell both constructions bracket the point estimate and
agree on the reading).

\begin{table}[H]
\centering
\caption{Both 95\% interval constructions for the ten Tier-3 cells.
Clustered: case-clustered percentile bootstrap ($B=10^4$). Run-level:
coupled bootstrap over test runs ($B=10^4$). Both bracket the point
estimate in every cell.}
\label{tab:ci}
\begin{tabular}{llccc}
\toprule
Model & Prompt & $\phi$ & $\phi$ CI (clustered) & $\phi$ CI (run-level) \\
\midrule
\texttt{qwen3.5-9b-ctx4k} & GPQA-ZS & 1.003 & [1.000, 1.007] & [0.969, 1.040] \\
\texttt{gemma4-26b} & GPQA-ZS & 1.002 & [1.000, 1.004] & [0.976, 1.028] \\
\texttt{qwen3.8-27b} & GPQA-ZS & 1.011 & [1.006, 1.017] & [0.975, 1.049] \\
\texttt{gemma4-31b} & GPQA-ZS & 1.001 & [1.000, 1.002] & [0.981, 1.020] \\
\texttt{qwen3.5-9b} & AIME-ZS & 1.056 & [1.045, 1.069] & [0.980, 1.141] \\
\texttt{gemma4-26b} & AIME-ZS & 1.018 & [1.014, 1.024] & [0.951, 1.089] \\
\texttt{qwen3.8-27b} & AIME-ZS & 1.052 & [1.042, 1.064] & [0.980, 1.131] \\
\texttt{gemma4-31b} & AIME-ZS & 1.013 & [1.009, 1.018] & [0.955, 1.075] \\
\texttt{qwen3.5-122b} & AIME-ZS & 1.041 & [1.031, 1.054] & [0.951, 1.141] \\
\texttt{qwen3.5-122b} & GPQA-ZS & 1.004 & [1.000, 1.008] & [0.966, 1.044] \\
\bottomrule
\end{tabular}
\end{table}

\section{Second model family (Qwen3.5-9B)}
\label{app:qwen}

This appendix is a precursor to the Tier-3 replication of
Section~\ref{sec:res-tier3}, which supersedes it: earlier sampling of
Qwen3.5-9B in this project (\texttt{ctx4k}) is re-used here
on two multiple-choice benchmarks with a different
protocol: MMLU ($C=4$, $K=16$, $n=500$ questions) and MMLU-Pro ($C=10$,
$K=32$, $n=70$). This arm predates the leak-free rival null, so it
reports the uniform per-question null $\rho=\agreementIndex_{\mathrm{iid,perq}}/
\agreementIndex_{\mathrm{emp}}$; on GPT-4.1 $\agreementIndex_{\mathrm{rival}}\ge
\agreementIndex_{\mathrm{iid}}$ holds in all eight cells, but the analogous
ordering is \emph{unverified} for Qwen3.5-9B (no rival null was computed
there: the Qwen collection stores one run per question, so a
leave-one-out per-case preference is undefined on it), so $\rho$ is read
as a same-metric point of contact, not as a
certified bound in the other direction. Two facts replicate. First, the mechanical coverage is of
the same order on a
different family: $\rho=0.556$ (MMLU) and $\rho=0.293$ (MMLU-Pro),
bracketing the GPT-4.1 uniform range $[0.340,0.546]$. Second, the coverage
\emph{declines} in the same direction as difficulty and/or answer-space
openness increase: within GPT-4.1 the rival share falls from
$\ge0.806$ (GPQA) to $\le0.781$ (AIME), a change in which answer-space
openness and difficulty are confounded, and within Qwen3.5-9B the
uniform coverage falls from $0.556$ (MMLU, $p=0.74$) to $0.293$ (MMLU-Pro,
$p=0.56$, also larger $C$ and $K$). The two arms are not merged into one table: $K$, benchmark, prompt, and sampling pipeline all differ, so only
this coarse, direction-consistent comparison is warranted. The Qwen arm
is committed native evidence (\texttt{analysis/llm\_selfconsistency.py},
\texttt{results/anchoring\_llm\_selfconsistency\_report.json}).

\section{Preference-estimate shrinkage}
\label{app:shrink}

The per-case preference $\hat q_i$ is estimated from the other runs of a
case, so on open-domain cases it is a noisy estimate. To bound the
influence of this estimation noise, the rival null is rerun with the
estimated preference shrunk toward uniform over its support,
$q_i(\lambda)=\lambda\hat q_i+(1-\lambda)\cdot\mathrm{uniform}$, for
$\lambda\in\{1,0.5,0\}$, where the support at $\lambda=0$ is the set of
answer labels that appear in the case's \emph{other} runs after
subtracting the test run's counts and dropping the ground truth and
\texttt{\_UNPARSEABLE\_}. (Settings: $n_{\mathrm{sim}}=2\times10^4$,
seed 0, $\texttt{--min-wrong 1}$; Monte Carlo SEs are reported per cell
in the evidence files \texttt{results/kappa\_rival\_shrink\_lam*.json}.)

\emph{Why $\lambda=0$ is not directly comparable to $\agreementIndex_{\mathrm{iid}}$.}
The uniform null draws over $C-1$ options with $C$ the \emph{per-run}
mean distinct-answer count, whereas the rival null at $\lambda=0$ draws
over the per-\emph{case} wrong-label support. These option spaces differ
materially on open-domain cells: across the AIME cells the mean rival
support is $19.5$--$69.4$ labels (median $17$--$57$, max $250$) against
$C=9$--$20$ (\texttt{results/kappa\_support\_audit.json}). A uniform
distribution over a larger option set dilutes the plurality, so
$\agreementIndex_{\mathrm{rival}}(\lambda=0)<\agreementIndex_{\mathrm{iid}}$ on AIME is
the expected direction, not an inconsistency: the two references are
uniform on \emph{different} option vocabularies, and Jensen's inequality
argument applies only when the two supports coincide (as they roughly do
on GPQA, where the mean rival support is $1.7$--$2.6$ against $C=4$).
The comparison of $\phi(\lambda=0)$ against $\rho$ in
\Cref{tab:shrink} is therefore directional evidence about the preference estimate, not a
 like-for-like benchmark comparison; the only like-for-like comparison is
across $\lambda$ values within a column, which uses the identical support
at every $\lambda$.

\begin{table}[H]
\centering
\caption{Mechanical coverage $\phi$ under preference-estimate shrinkage.}
\label{tab:shrink}
\begin{tabular}{lccc}
\toprule
$\lambda$ (shrinkage) & $1$ & $0.5$ & $0$ \\
\midrule
GPQA $\phi$ range & $[0.808,0.927]$ & $[0.681,0.840]$ & $[0.593,0.781]$ \\
AIME $\phi$ range & $[0.584,0.777]$ & $[0.386,0.479]$ & $[0.206,0.292]$ \\
\bottomrule
\end{tabular}
\end{table}

All three columns share the same $\lambda$-invariant denominator
$\agreementIndex_{\mathrm{emp}}^{(t)}$ (verified bit-identical across the three
runs). Two mechanical notes. First, the rival numerator averages over
test runs with a defined simulated expectation (a test run with no
simulated wrong plurality contributes no draw): at
$n_{\mathrm{sim}}=2\times10^4$ this is $\ge76\%$ of test runs per cell ($\ge87\%$ in six of eight cells),
and at the main-table setting ($10^5$) the draw coverage is $78$--$96\%$ across cells (the shortfall reflects high-accuracy cases whose simulation never produces a wrong plurality). Second, the $\lambda=1$ column reproduces
Table~\ref{tab:kappa} within Monte Carlo error
($n_{\mathrm{sim}}=2\times10^4$ here vs.\ $10^5$ there; differences
$\le0.009$).

Two conclusions. First, on GPQA the mechanical coverage stays above $0.5$ even when the estimated preference is treated as pure noise
($\lambda=0$): the multiple-choice result does not ride on the
preference estimate. Second, on AIME $\phi$ declines monotonically with
$\lambda$ within a fixed support, so the residual $1-\phi$ is
\emph{largest} at $\lambda=0$: the headline AIME residual (at
$\lambda=1$) is, in this direction, conservative rather than inflated
by estimation noise. The benchmark split thus survives shrinkage in
direction at every level, with only its magnitude attenuated. Monte
Carlo SEs of these $\agreementIndex$ values are $\le4\times10^{-3}$ $\agreementIndex$ units (reported per cell in the evidence files). A draw-coverage convention is disclosed: test runs whose simulation never produced a wrong plurality contribute to the denominator $\agreementIndex_{\mathrm{emp}}^{(t)}$ but not to the numerator. Counting them as zero in the numerator (the most conservative alternative) lowers $\phi$ by $0.03$--$0.17$ per cell (largest, $0.17$, on the
lowest-coverage cell at $76\%$ draw coverage;
\texttt{results/kappa\_shrink\_conservative.json}), and the benchmark
direction survives: GPQA $\phi$ stays $0.75$--$0.84$ against AIME
$0.51$--$0.71$.

\section{Calibrated explanatory null: run-level preference heterogeneity}
\label{app:dispersion}

The rival null fixes each case's wrong-label preference $\hat q_i$ and
treats every run of the case as an i.i.d.\ draw from it. A stronger
reference also grants the case \emph{run-level preference
heterogeneity}: before drawing its $K$ votes, each simulated run draws
its own preference $q^*\sim\mathrm{Dirichlet}(\alpha_{\mathrm{disp}}\,\hat q_i)$ and
then draws its wrong votes from $q^*$. $\alpha\to\infty$ recovers the
multinomial rival; small $\alpha$ means the runs of a case disagree
strongly about which wrong answers are attractive. One $\alpha$ per cell is fitted by moment-matching the simulated cross-run variance of the
plurality share to the observed within-case cross-run variance of the
plurality share, fitted on a random held-out half of the eligible cases and evaluated on the other half (cases with $\ge3$ wrong runs;
$n_{\mathrm{sim}}=2\times10^3$, seed 0;
\texttt{results/kappa\_rival\_dispersion.json}). The fit and
evaluation halves contain $8$--$83$ eligible cases per cell
(\texttt{results/kappa\_rival\_dispersion.json}). The denominator of
$\phi_{\mathrm{dm}}$ is the empirical index over the
dispersion-eligible population (cases with $\ge3$ wrong runs), not the
test-subset index $\agreementIndex_{\mathrm{emp}}^{(t)}$. \Cref{tab:dispersion}
reports the fitted $\alpha$ per cell with the observed and simulated
cross-run variance and the resulting coverage $\phi_{\mathrm{dm}}$.

\begin{table}[H]
\centering
\caption{Run-level preference heterogeneity: fitted Dirichlet $\alpha$ and coverage of the calibrated null $\phi_{\mathrm{dm}}$.}
\label{tab:dispersion}
\begin{tabular}{llcccc}
\toprule
Model & Prompt & $\alpha$ fit & $\mathrm{var}_{\mathrm{obs}}$ &
$\mathrm{var}_{\mathrm{DM}}$ & $\phi_{\mathrm{dm}}$ \\
\midrule
\texttt{4.1} & AIME-ZS & 1.5 & 0.026 & 0.026 & 1.38 \\
\texttt{4.1} & GPQA-ZS & 1.5 & 0.012 & 0.003 & 1.01 \\
\texttt{4.1-mini} & AIME-CoT & 0.5 & 0.038 & 0.010 & 1.68 \\
\texttt{4.1-mini} & AIME-ZS & 0.5 & 0.055 & 0.021 & 1.74 \\
\texttt{4.1-mini} & GPQA-CoT & 25 & 0.016 & 0.004 & 0.89 \\
\texttt{4.1-mini} & GPQA-ZS & 50 & 0.021 & 0.004 & 0.85 \\
\texttt{4.1-nano} & AIME-ZS & 1.0 & 0.048 & 0.027 & 2.10 \\
\texttt{4.1-nano} & GPQA-ZS & 1.0 & 0.022 & 0.008 & 1.01 \\
\bottomrule
\end{tabular}
\end{table}

Three results. First, the observed run-level dispersion of the plurality
share is $2$--$3\times$ larger on AIME than on GPQA, i.e., open-domain
runs of the same case show substantially greater cross-run dispersion in which wrong answers are
attractive; this dispersion itself is the empirical signature of
cross-run structure beyond a fixed per-case marginal. Second, on GPQA
the fitted $\alpha$ spans $1.0$--$50$ (moderate; two cells sit at $1.0$--$1.5$) and
$\phi_{\mathrm{dm}}$ sits near the multinomial rival
($0.85$--$1.01$): the multiple-choice conclusion is unchanged. Third,
on AIME the fitted $\alpha$ is $0.5$--$1.5$ and
$\phi_{\mathrm{dm}}=1.38$--$2.10$: a reference that grants the observed
run-level heterogeneity \emph{more than} reproduces the empirical
agreement index. Two caveats apply. First, the dispersion
parameter is fitted on a random held-out half of the eligible cases and
evaluated on the held-out half, so the parameter never touches the data
it explains, an out-of-sample evaluation on held-out cases rather than an in-sample calibrated fit.
Second, the Dirichlet form under-disperses the observed spread in every cell, worse on GPQA ($2.7$--$6.1\times$) than on AIME ($1.0$--$3.7\times$), so the fitted $\alpha$ is a lower envelope and the $\phi_{\mathrm{dm}}$ values under-grant heterogeneity. (The ratio ranges are computed from the unrounded variances (the table above shows them to three decimals), so they do not always reproduce from the displayed entries; each ratio is the observed variance divided by the fitted-$\alpha$ variance of the same cell. The fit is a moment match on the grid $\{0.2,0.5,1.0,1.5,3,6,12,25,50,\infty\}$; no fitted value sits at the grid floor. Across three independent fit/eval split seeds, the ranges are stable: GPQA $\phi_{\mathrm{dm}}\in[0.76,1.05]$, AIME $\in[1.36,2.10]$.) The per-seed spread of each simulated $\mathrm{var}_{\mathrm{DM}}$ is small (relative SE $\le7\%$ across the three seeds at $n_{\mathrm{sim}}=2\times10^3$), so the dispersion null is a robust lower-envelope diagnostic rather than a precision target.
The preference-unexplained
residual on AIME is therefore more than absorbed by this calibrated
run-level preference-heterogeneity null: on open-domain tasks the runs of a case do
not share one attractive wrong answer; they differ in which wrong
answers attract them, a correlated-error structure at the level of
run-to-run preference variation. The residual $\delta$ should
accordingly be read as a signature \emph{consistent with this channel},
not as an identified mechanism.

\section{Backfire on hard questions and a low empirical high-agreement ceiling}
\label{app:backfire}

\begin{figure}[H]
\centering
\includegraphics[width=0.95\textwidth]{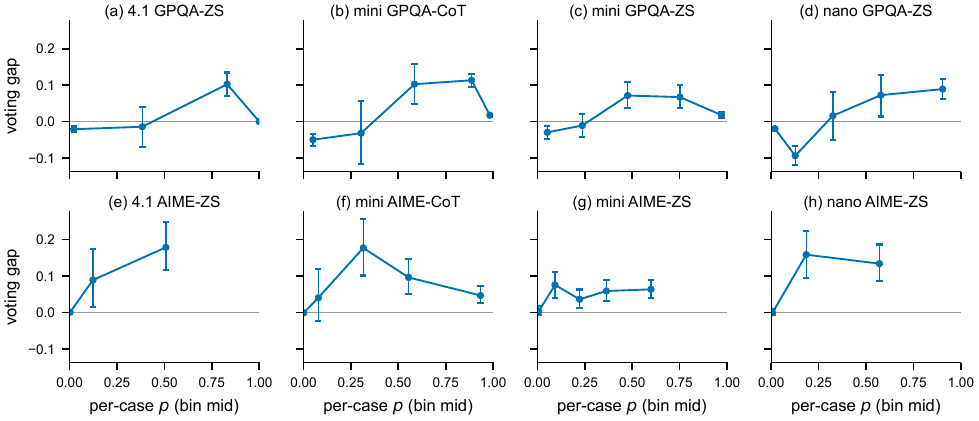}
\caption{Voting gap (consensus $-$ single-sample accuracy) against
per-case difficulty (bins at the 20/40/60/80 percentiles of $p$), with
coupled case-level
bootstrap intervals on each bin. Panels (a)--(h): the eight
model--benchmark--prompt cells of Table~\ref{tab:samples}. Negative
values on the hardest bins
are the backfire signature}
\label{fig:backfire}
\end{figure}

The backfire phenomenon of \cite{bahuguna2026when} is reproduced in the
difficulty-binned view: cases are split into five bins at the
20/40/60/80 percentiles of per-case accuracy (ties fall into the upper
bin, so bins are not exactly equal-frequency when many cases share an
accuracy), and the voting gap (consensus accuracy minus single-sample
accuracy) is computed per bin. Majority voting helps
on easy questions but is neutral or
harmful on hard ones. The most negative voting gaps occur on the hardest
bins; three representative cells:
\begin{itemize}
\item \texttt{gpt-4.1-nano} GPQA-ZS, cases with per-case $p\approx0.13$:
  gap $= -0.09$, coupled case-level bootstrap CI $[-0.12, -0.07]$ ($n=37$ cases).
\item \texttt{gpt-4.1-mini} GPQA-CoT, hardest bin ($p\approx0.05$):
  gap $= -0.05$, coupled case-level bootstrap CI $[-0.07, -0.03]$ ($n=36$ cases).
\item \texttt{gpt-4.1-mini} GPQA-ZS ($p\approx0.24$): gap $= -0.01$,
  coupled CI $[-0.04, 0.02]$ (crosses zero; $n=39$ cases).
\end{itemize}
These intervals come from a coupled case-level bootstrap ($B=10^4$; consensus and single-sample accuracy of a sampled case move together), the appropriate paired unit for the gap. The bins are defined on the full data and evaluation occurs within those same bins (no independent holdout for bin definition); the intervals therefore quantify within-bin sampling variability, not the uncertainty of the binning rule itself. The bins are fixed once before resampling, so each bootstrap draw resamples cases \emph{within} the same fixed bin; because a bin is narrow in per-case single-sample accuracy, the paired gap varies across cases mainly through the coupling of the two accuracies rather than through residual within-bin dispersion in $p_i$. This is what makes the intervals tight (the hardest-bin CI spans only $0.05$) despite the modest per-bin sample sizes. The two hardest-bin gaps exclude zero, while the third crosses
zero: the difficulty-binned backfire has its CI excluding zero on the
two hardest bins and is directionally consistent with
\cite{bahuguna2026when}, reported as a replication of the pattern,
not as an
independent precision claim.
The global gap is positive because easy cases dominate the sample; the difficulty-binned (not pooled) gap is reported so the backfire is visible
and properly scoped. \Cref{fig:backfire} shows the full binned picture
for all eight cells.

A second, complementary ceiling signature appears in the agreement-binned view. ``Ceiling'' is used in the \emph{empirical} sense
only (the observed accuracy of the highest-agreement bin), not as a
theoretical upper bound on achievable accuracy. Runs are split into
five equal-quantile
(quintile) bins on the run's self-consistency $\alpha$ (the same
equal-quantile rule as the difficulty bins above); the
``highest-agreement bin'' is the top quintile.
\Cref{tab:ceiling} reports the consensus accuracy
within the highest-agreement bin of each cell, alongside the cell base
rate $p$. Two facts hold. First, the ceiling is a substantial lift over
the base rate: from $1.2\times$ (\texttt{4.1} GPQA-ZS, $0.59$ vs.\
$p=0.47$) to $3.6\times$ (\texttt{4.1-nano} AIME-ZS, $0.59$ vs.\
$p=0.16$), so high agreement \emph{does} carry predictive signal. Second,
the ceiling is nevertheless far below $1.0$: it reaches only
$0.42$--$0.83$, and for the weaker models
(\texttt{gpt-4.1-nano}, \texttt{gpt-4.1} on AIME) it is as low as
$0.42$--$0.59$. High self-consistency is therefore informative but far
from reliable, consistent with the decomposition of
Section~\ref{sec:res-kappa} (where open-domain cells retain a substantial
preference-unexplained residual).

\begin{table}[H]
\centering
\caption{Consensus accuracy in the highest-agreement bin (top quintile
of run self-consistency $\alpha$), with the cell
base rate $p$ (single-sample accuracy) and 95\% Wilson CIs. Because many
runs sit at $\alpha=1.0$, the top quintile can contain more than 20\% of
runs and for three GPQA cells (\texttt{4.1} GPQA-ZS, \texttt{mini} GPQA-CoT, \texttt{mini} GPQA-ZS) consists of $\alpha=1.0$ runs alone.
``Ceiling'' is how far high agreement reaches toward a correct consensus: it is a clear lift over
$p$ but stays far below $1.0$. $n$ is the number of runs
falling in that bin (a run is one $K=50$ sampling pass;
Section~\ref{sec:data}).}
\label{tab:ceiling}
\begin{tabular}{llcccc}
\toprule
Model & Prompt & ceiling & CI & $p$ & $n$ (runs) \\
\midrule
\texttt{4.1} & AIME-ZS & 0.42 & [0.32, 0.52] & 0.12 & 93 \\
\texttt{4.1} & GPQA-ZS & 0.59 & [0.52, 0.65] & 0.47 & 227 \\
\texttt{4.1-mini} & AIME-CoT & 0.83 & [0.74, 0.89] & 0.38 & 95 \\
\texttt{4.1-mini} & AIME-ZS & 0.74 & [0.69, 0.79] & 0.26 & 290 \\
\texttt{4.1-mini} & GPQA-CoT & 0.83 & [0.74, 0.89] & 0.55 & 92 \\
\texttt{4.1-mini} & GPQA-ZS & 0.74 & [0.69, 0.78] & 0.49 & 445 \\
\texttt{4.1-nano} & AIME-ZS & 0.59 & [0.48, 0.68] & 0.16 & 92 \\
\texttt{4.1-nano} & GPQA-ZS & 0.55 & [0.45, 0.65] & 0.39 & 93 \\
\bottomrule
\end{tabular}
\end{table}

\section{Champion fragility: the winner itself is not stable}
\label{app:fragility}

The mechanical coverage $\phi$ and the low ceiling both concern the
\emph{strength} of
a single consensus answer. A complementary question is \emph{stability}:
given two $K=50$ sampling runs of the same case and prompt under nominally identical settings,
does the plurality winner agree? Ding's per-run data marks each run with
one of two sampling conditions ($a$/$b$); for \texttt{gpt-4.1-mini}, 353
of the 394 cases carry at least one run under each condition (178 GPQA-ZS
and 175 AIME-ZS; Section~\ref{sec:data}), and each such case's two conditions are paired,
taking the first run of each condition when a case has several. One structural fact shapes the interpretation: every \texttt{mini}-ZS pair is \emph{cross-axis} (condition $b$ exists only under axis $A$ and condition $a$ only under axes $B$/$C$), so this section measures cross-batch winner stability, not same-parameter replication; the AIME flip rate in particular confounds axis (batch) differences with sampling randomness. As a sensitivity, pairing only $A/b$ against $B/a$ (excluding the lower-accuracy $C/a$ arm; \texttt{results/champion\_axis\_sensitivity.json}) changes the AIME flip rate from $0.82$ to $0.80$ ($n=155$) and leaves the GPQA rate at $0.41$ ($n=160$): the qualitative finding is unchanged. The first-run choice is immaterial: recomputing with the last
run or a random run changes the overall flip rate by at most $0.06$ (\texttt{results/champion\_tierun\_sensitivity.json}).
The \emph{champion flip rate} is defined as the
fraction of pairs on which the two runs select different plurality
winners; this requires no new sampling and no assumptions beyond
the pairing.

\Cref{tab:fragility} reports the flip rate overall and binned by the
self-consistency $\alpha$ of the run from condition $a$. Two findings stand out. First,
the flip rate is substantial: $0.40$ on GPQA and $0.82$ on AIME, where
open-ended answers give many distinct candidates that a $K=50$ sample
cannot reliably separate. The two benchmark-level rates are therefore
not directly comparable as a measure of fragility per se: the AIME rate
conflates winner instability with the larger candidate set. Second, high agreement only weakly stabilizes
the winner: on GPQA the flip rate falls from $0.64$ in the lowest bin to
$0.17$ at $\alpha\geq0.98$, but it does not vanish; even nearly unanimous
runs pick a different champion $17\%$ of the time. Champion fragility is
thus a complementary signature of winner instability under the
released cross-condition protocol, reported before shared error is
considered.

\begin{table}[H]
\centering
\caption{Champion flip rate across the two sampling conditions of the
same case and prompt (\texttt{gpt-4.1-mini}). ``Flip rate'' is the fraction of
pairs where the plurality winner differs between the case's two sampling
conditions; 95\% bootstrap CIs in parentheses; $\alpha$ bins use the
self-consistency of the run from condition $a$, split into three
equal-quantile bins (hence the reported boundaries).}
\label{tab:fragility}
\begin{tabular}{llccc}
\toprule
Benchmark & Bin & flip rate & $n$ (pairs) \\
\midrule
GPQA-ZS & overall & 0.40 (0.33, 0.47) & 178 \\
GPQA-ZS & $\alpha\in[0.28,0.72)$ & 0.64 (0.52, 0.76) & 58 \\
GPQA-ZS & $\alpha\in[0.72,0.98)$ & 0.44 (0.30, 0.58) & 50 \\
GPQA-ZS & $\alpha\geq0.98$ & 0.17 (0.09, 0.26) & 70 \\
\midrule
AIME-ZS & overall & 0.82 (0.77, 0.88) & 175 \\
AIME-ZS & $\alpha\in[0.06,0.28)$ & 0.91 (0.81, 0.98) & 54 \\
AIME-ZS & $\alpha\in[0.28,0.84)$ & 0.79 (0.69, 0.89) & 61 \\
AIME-ZS & $\alpha\geq0.84$ & 0.78 (0.68, 0.88) & 60 \\
\bottomrule
\end{tabular}
\end{table}

\section{Jensen gap: soft agreement adds nothing over hard}
\label{app:jensen}

A standard response to winner fragility is to replace hard majority
voting with soft, probability-weighted aggregation. In this setting only
the aggregate answer counts of each run are available (no per-sample
logits), so the faithful analogue is a comparison of two scoring rules on
the \emph{same} hard winner: hard confidence $H=\max_c p_c$ (the self-consistency $\alpha$) versus soft confidence $S=\sum_c p_c^2$, the
probability that two independent draws agree. The \emph{Jensen gap}
$J = S - H^2 \ge 0$ isolates the agreement mass that lies beyond the
plurality answer.

\Cref{tab:jensen} reports AUROC against plurality correctness and the
mean gap. 
\begin{table}[H]
\centering
\caption{Soft vs.\ hard agreement as confidence signals. $H=\max_c p_c$
is self-consistency; $S=\sum_c p_c^2$; $J=S-H^2$. AUROC predicts whether
the plurality answer is correct. Cells aggregate all runs of a model on a
benchmark (across prompts); pooled row is global. $J$ split by plurality
correctness is shown in the last two columns. Cell membership is resolved
through the $(\text{axis},\text{condition})$ map of Section~\ref{sec:data}.}
\label{tab:jensen}
\begin{tabular}{llccccc}
\toprule
Model & Benchmark & AUROC($H$) & AUROC($S$) & mean $J$ & $J$, wrong & $J$, correct \\
\midrule
\texttt{4.1} & AIME & 0.739 & 0.738 & 0.052 & 0.055 & 0.036 \\
\texttt{4.1} & GPQA & 0.607 & 0.608 & 0.027 & 0.034 & 0.020 \\
\texttt{4.1-mini} & AIME & 0.847 & 0.837 & 0.041 & 0.054 & 0.016 \\
\texttt{4.1-mini} & GPQA & 0.678 & 0.677 & 0.047 & 0.062 & 0.033 \\
\texttt{4.1-nano} & AIME & 0.820 & 0.806 & 0.041 & 0.046 & 0.025 \\
\texttt{4.1-nano} & GPQA & 0.621 & 0.630 & 0.055 & 0.062 & 0.045 \\
\midrule
\multicolumn{2}{c}{Pooled} & 0.770 & 0.769 & 0.044 & 0.055 & 0.027 \\
\bottomrule
\end{tabular}
\end{table}
Two findings matter. First, no appreciable AUROC improvement is observed from the distributional score $S$ over the plurality concentration $H$
($0.769$ vs.\ $0.770$
pooled, within $0.02$ on every cell; per-cell $0.61$--$0.85$), so
agreement carries real but bounded predictive signal, consistent with
the ceiling analysis. This is a
null
result: soft weighting does not beat hard majority, so it is reported as
an understanding contribution, not
a new voting method.
Second, the Jensen gap is diagnostic: it is roughly twice as large on
runs whose plurality answer is \emph{wrong} (mean $0.055$) than on
correct ones ($0.027$). When the winner is wrong, the empirical
distribution is more spread across candidates. This unifies the previous
signatures: wrong runs show higher distributional dispersion than
correct ones, and the additional distributional mass is associated
with wrong plurality outcomes; the available scores do not improve on
plurality concentration.

\section*{Author contributions}
\label{app:contrib}

Conceptualization and methodology: L.\ Zhang; software, formal analysis,
investigation, data curation, and visualization: M.\ Tang and L.\ Zhang;
writing (original draft preparation): L.\ Zhang and M.\ Tang; writing
(review and editing): C.\ Long, X.\ Tang, X.\ Luo, and L.\ Zhang;
supervision: L.\ Zhang, C.\ Long, and X.\ Tang; resources and funding
acquisition: C.\ Long, X.\ Tang, and X.\ Luo.

\end{document}